\documentclass[a4paper,journal,compsoc]{IEEEtran}

\usepackage{algpseudocode}
\usepackage{color}
\usepackage{csquotes}
\usepackage{subfigure} 
\usepackage{blkarray}
\usepackage{algorithm}
\usepackage{array}
\usepackage{subfigure}
\usepackage{graphicx}         
\usepackage{verbatim}         
\usepackage{amssymb}          
\usepackage{amsmath}          
\usepackage{arydshln}

\usepackage{amsthm} 

\newtheorem{lem}{Lemma}

\newtheorem{prop}{Proposition}

\theoremstyle{definition}
\newtheorem{definition}{Definition}

\newtheorem{example}{Example}

\usepackage{calc}
\usepackage[noadjust]{cite}
\usepackage{tikz}
\usepackage{pgfplots} 
\usetikzlibrary{decorations.markings}
\tikzstyle{vertex}=[circle, draw, inner sep=0pt, minimum size=6pt]

\usepackage{array}

\makeatletter
\newcommand{\thickhline}{%
    \noalign {\ifnum 0=`}\fi \hrule height 1pt
    \futurelet \reserved@a \@xhline
}
\newcolumntype{"}{@{\hskip\tabcolsep\vrule width 1pt\hskip\tabcolsep}}
\makeatother

\newcommand{\ent}[1]{H(#1)}

\newcommand{\mutinf}[1]{I(#1)}

\newcommand{\loss}[2][\empty]{\ifthenelse{\equal{#1}{\empty}}{L(#2)}{L_{#1}(#2)}}
\newcommand{\lossrate}[2][\empty]{\ifthenelse{\equal{#1}{\empty}}{L(\mathbf{#2})}{L_{\mathbf{#1}}(\mathbf{#2})}}

\newcommand{\relLoss}[2][\empty]{\ifthenelse{\equal{#1}{\empty}}{l(#2)}{l_{#1}(#2)}}

\DeclareMathOperator*{\argmin}{arg\,min}

\newcommand{\dom}[1]{\mathcal{#1}}

\newcommand{\Wvec}{\mathbf{W}}

\newcommand{\Amat}{\mathbf{A}}

\newcommand{\numitermax}{\text{\#iter}_{\max}}
\newcommand{\numiter}{\text{\#iter}}
\newcommand{\toler}{\text{tol}}
\newcommand{\Rvec}{\mathbf{R}}

\newcommand{\rev}[1]{#1}

\begin{document}

\title{Co-Clustering via Information-Theoretic Markov Aggregation} 

\author{Clemens Bl\"ochl, %
Rana Ali Amjad,~\IEEEmembership{Student Member,~IEEE,} %
and Bernhard C. Geiger,~\IEEEmembership{Member,~IEEE}%
\IEEEcompsocitemizethanks{\IEEEcompsocthanksitem Clemens Bl\"ochl was with the Institute for Communications Engineering, Technical University of Munich, Germany and is now with Rohde \& Schwarz GmbH \& Co.\ KG, Munich, Germany.\IEEEcompsocthanksitem Rana Ali Amjad is with the Institute for Communications Engineering, Technical University of Munich, Germany. Email: ranaali.amjad@tum.de\IEEEcompsocthanksitem Bernhard C. Geiger was with the Signal Processing and Speech Communication Laboratory, Graz University of Technology, Austria and is now with KNOW-CENTER Gmbh, Graz, Austria. Email: geiger@ieee.org} \thanks{The authors contributed equally to this work.}
}

\IEEEtitleabstractindextext{
\begin{abstract}
We present an information-theoretic cost function for co-clustering, i.e., for simultaneous clustering of two sets based on similarities between their elements. By constructing a simple random walk on the corresponding bipartite graph, our cost function is derived from a recently proposed generalized framework for information-theoretic Markov chain aggregation. The goal of our cost function is to minimize \emph{relevant information loss}, hence it connects to the information bottleneck formalism. Moreover, via the connection to Markov aggregation, our cost function is not \emph{ad hoc}, but inherits its justification from the operational qualities associated with the corresponding Markov aggregation problem. We furthermore show that, for appropriate parameter settings, our cost function is identical to well-known approaches from the literature, such as Information-Theoretic Co-Clustering of Dhillon et al. Hence, understanding the influence of this parameter admits a deeper understanding of the relationship between previously proposed information-theoretic cost functions. We highlight some strengths and weaknesses of the cost function for different parameters. We also illustrate the performance of our cost function, optimized with a simple sequential heuristic, on several synthetic and real-world data sets, including the Newsgroup20 and the MovieLens100k data sets.
\end{abstract}
}

\maketitle

\section{Introduction and Outline}
Co-clustering is the task of the simultaneous clustering of two sets,
typically represented by rows and columns of a data matrix. Aside from being a clustering problem in its own right, co-clustering is also applied for clustering only one dimension of the data matrix. In these scenarios, co-clustering is an implicit method for feature clustering and provides an alternative to feature selection with, purportedly, increased robustness to noisy data~\cite{Slonim_DoubleClustering,Dhillon_ITClustering,Wang_IBCC}. 

A popular approach to co-clustering employs information-theoretic cost functions and is based on transforming the data matrix into a probabilistic description of the two sets and their relationship. For example, if the entries in the data matrix are all nonnegative, one can normalize the data matrix to obtain a joint probability distribution of two random variables taking values in the two sets. This approach has been taken by, e.g., Slonim et al.~\cite{Slonim_DoubleClustering}, Bekkerman et al.~\cite{Bekkerman_MultiClustering}, El-Yaniv and Souroujon~\cite{El-Yaniv_ITDC}, and Dhillon et al.~\cite{Dhillon_ITClustering} (see also Section~\ref{sec:relatedwork}). A different approach to co-clustering is to identify the data matrix with the weight matrix of a bipartite graph and subsequently apply graph partitioning methods to cluster the rows and columns of the data matrix. This approach has been taken by, e.g., Dhillon~\cite{Dhillon_Graph}, Labiod and Nadif~\cite{Labiod_Modularity}, and Ailem et al.~\cite{Ailem_Modularity}. Other popular approaches are model-based (e.g., latent block models as in~\cite{PLBM_Nadif} and the references therein) or based on nonnegative matrix factorization (e.g.,~\cite[Sec.~4.4]{Sra_NMF}).

In this work, we combine ideas from the graph-based and the information-theoretic approaches. Specifically, we use the data matrix to define a simple random walk on a bipartite graph, i.e., a first-order, stationary Markov chain. Clustering this bipartite graph (i.e., co-clustering) thus becomes equivalent to clustering the state space of a Markov chain (i.e., Markov aggregation, cf.~Section~\ref{sec:MarkovAggregation}). This, in turn, allows us to transfer the information-theoretic cost function from the latter problem to the former. The thus presented cost function, parameterized by a single parameter $\beta$, derives its justification from the corresponding Markov aggregation problem. This justification is further inherited to other information-theoretic cost functions previously proposed in the literature~\cite{Slonim_DoubleClustering,Slonim_DocClustering,Dhillon_ITClustering,Bekkerman_MultiClustering,Wang_IBCC}, which we obtain as special cases for appropriate choices of $\beta$. 

\rev{In several examples we discuss weaknesses inherent in the cost function for certain values (or value ranges) of $\beta$.} We also present a simple sequential heuristic to optimize our cost function and analyze the influence of the choice of $\beta$ on the co-clustering performance. \rev{For the synthetic data sets, we confirm that co-clustering outperforms one-sided clustering if the data matrix is noisy or if there is strong intra-cluster coupling}. For the Newsgroup20 data set we observed that performance is insensitive to $\beta$ as long as the number of word clusters is sufficiently large. Performance drops for few word clusters, a fact for which we provide a theoretical explanation. The parameter $\beta$ has a somewhat stronger influence on the performance on the MovieLens100k data set, for which we obtained movie clusters largely consistent with genres. Finally, for the Southern Women Event Participation Dataset, our results are remarkably similar to the reference co-clusterings from~\cite{Barber_Modularity,Doreian_Blockmodeling}.

In summary, our contribution is threefold:
\begin{enumerate}
 \item We provide a generalized framework for information-theoretic co-clustering via connecting it with Markov aggregation. The cost function, parameterized with a single parameter and connected with the information bottleneck formalism, is justified by well-defined operational goals of the Markov aggregation problem (Sections~\ref{sec:MarkovAggregation} \&~\ref{sec:CostFunctions}).
 \item Our generalized framework contains previously proposed information-theoretic cost functions as special cases (Section~\ref{sec:specialAlgos}). Since the parameter of our cost function has an intuitive meaning, our framework leads to a deeper understanding of the previously proposed approaches. This understanding is further developed by pointing at the \rev{strengths and limitations of information-theoretic cost functions for co-clustering with the help of examples and experiments on synthetic datasets (Section~\ref{sec:pathological}). We also discuss the influence of the single parameter on the co-clustering results and present general guidelines for setting this parameter depending on the characteristics of the dataset.}
 \item \rev{We perform experiments (Section~\ref{sec:experiments}) with real-world datasets. Varying the parameter allows us to compare our results to those obtained via cost functions previously proposed in the literature}. 
\end{enumerate}
We do not address the important issues of choosing the number of clusters, nor do we design sophisticated optimization heuristics and/or initialization procedures; essentially, most heuristics proposed for previous cost functions such as in \cite{Dhillon_ITClustering,Slonim_DocClustering } can be adapted to our framework.

The fact that our cost function contains previously proposed cost functions as special cases allows us to compare them \emph{fairly}, i.e., with the same initialization steps and the same optimization heuristic. For example, the insensitivity to $\beta$ in our experiments with the Newsgroup20 datasets provides a new perspective on the differences reported in~\cite{Dhillon_ITClustering,Wang_IBCC,Slonim_DoubleClustering,Bekkerman_MultiClustering}, suggesting that they are due to differences in optimization heuristics, preprocessing steps, or choice of data subsets rather than due to differences in the cost function. 

\textbf{Notation.}
Random variables (RVs) are denoted by upper case letters ($Z$), lower case letters ($z$) are reserved for realizations and constants, and calligraphic letters ($\mathcal{Z}$) are used for sets. We use bold upper case letters ($\mathbf{Z}$) to denote matrices. We assume that the reader is familiar with information-theoretic quantities. Specifically, the \emph{mutual information} between two RVs $Z$ and $S$ with finite alphabet and joint distribution $P_{Z,S}$ is denoted as $I(Z;S)$~\cite[eq.~(2.28)]{Cover_Information2}. Note further that $I(Z;S)=H(S)-H(S|Z)$, where $H(S)$ is the entropy of $S$ and where $H(S|Z)$ is the conditional entropy of $S$ given $Z$.

\section{Related Work}\label{sec:relatedwork}
\subsection{Information-Theoretic Co-Clustering Approaches}
Information-theoretic approaches to co-clustering require a probability distribution over the sets to be clustered, which we will denote as $\dom{X}$ and $\dom{Y}$. For example, if the data matrix $\Wvec$ is nonnegative, then one can normalize it such that its entries sum to one. One can thus define RVs $X$ and $Y$ over the sets $\dom{X}$ and $\dom{Y}$ that have a joint distribution $P_{X,Y}\propto \Wvec$.

One-sided clustering, i.e., clustering only the RV $X$ with a clustering function $\Phi$ such that information about $Y$ is preserved, was one of the main motivations behind the information bottleneck (IB) method~\cite{Tishby_InformationBottleneck}. Several algorithmic approaches have been proposed, including agglomerative~\cite{Slonim_Agglomerative} and sequential~\cite{Slonim_DocClustering} methods and a method reminiscent of k-means~\cite{Dhillon_Divisive} (the latter being equivalent to the fixed-point iterations in the original paper~\cite{Tishby_InformationBottleneck}).

An early information-theoretic approach to co-clustering was proposed by Slonim and Tishby~\cite{Slonim_DoubleClustering} and is based on the IB method~\cite{Tishby_InformationBottleneck}. There, the authors proposed first finding the clustering function $\Phi$ maximizing $I(\Phi(X);Y)$, and then, after fixing $\Phi$, finding the clustering function $\Psi$ that maximizes $I(\Phi(X);\Psi(Y))$. Their approach was improved later by El-Yaniv and Souroujon, who suggested iterating this procedure multiple times~\cite{El-Yaniv_ITDC}. Also based on the IB method is the work of Wang et al.~\cite{Wang_IBCC}. They used a multivariate extension of mutual information to compress ``input information'' -- captured by the mutual information terms $I(X;Y)$, $I(X;\Phi(X))$, and $I(Y;\Psi(Y))$ -- while preserving relevant information -- captured by the information shared between the clusters, $I(\Phi(X);\Psi(Y))$, and the predictive power of the clusters, $I(\Phi(X);Y)$ and $I(X;\Psi(Y))$.

In 2003, Dhillon et al. proposed a co-clustering algorithm simultaneously determining clustering functions $\Phi$ and $\Psi$ with the goal to maximize $I(\Phi(X);\Psi(Y))$~\cite{Dhillon_ITClustering}. They showed that the problem is equivalent to a constrained nonnegative matrix tri-factorization problem~\cite[Lemma~2.1]{Dhillon_ITClustering} with Kullback-Leibler divergence as cost function. (An iterative update rule for the entries of the three matrices is provided in~\cite[Sec.~4.4]{Sra_NMF}.) The work in~\cite{Dhillon_ITClustering} was generalized into various directions. On the one hand, Bekkerman et al. investigated simultaneous clustering of more than two sets in~\cite{Bekkerman_MultiClustering}. Rather than maximizing one of the multivariate extension of mutual information, the authors suggested maximizing the sum of mutual information terms between pairs of clusters; the pairs of clusters considered in the sum are determined by an undirected graph that has to be provided by the user. On the other hand, Banerjee et al. viewed co-clustering as a matrix approximation problem~\cite{Banerjee_Bregman}, of which the nonnegative matrix tri-factorization problem of~\cite[Lemma~2.1]{Dhillon_ITClustering} is a special case. Their generalized framework admits any Bregman divergence (e.g., Kullback-Leibler divergence or squared Euclidean distance) as cost function and several co-clustering schemes characterized by the type of summary statistic used to approximate the matrix.

Finally, Laclau et al. formulate the co-clustering problem as an optimal transport problem with entropic regularization~\cite{Laclau_OptimalTransport}. Their formulation also turns into a probability matrix approximation problem with Kullback-Leibler divergence as cost function, but 1) the order of original and approximate distribution is swapped compared to~\cite[Lemma~2.1]{Dhillon_ITClustering}, and 2) the approximate distribution is obtained differently. They proposed solving the co-clustering problem with the Sinkhorn-Knopp algorithm and suggested a heuristic to determine the number of clusters.

\rev{\subsection{Markov Aggregation and Lumpability}
Markov aggregation is the task of replacing a Markov chain $\{Z_t{:}\ t = 1, 2,\dots\}$ with a alphabet $\mathcal{Z}$ by a Markov chain with a smaller alphabet $\overline{\mathcal{Z}}$, sacrificing model accuracy for a reduction in model complexity. Aggregation is usually performed by partitioning (i.e., clustering) the alphabet $\mathcal{Z}$ and defining a Markov chain on the partitioned alphabet $\overline{\mathcal{Z}}$. Information-theoretic cost functions for Markov aggregation had been proposed in, e.g.,~\cite{Meyn_MarkovAggregation,GeigerEtAl_OptimalMarkovAggregation,Xu_Reduction} and were recently unified in~\cite{Amjad_GeneralizedMA}. More generally, aggregations of dynamical systems that are not necessarily Markov were discussed in~\cite{Wolpert_HighLevel}. In contrast to~\cite{Meyn_MarkovAggregation,GeigerEtAl_OptimalMarkovAggregation,Xu_Reduction,Amjad_GeneralizedMA}, the cost functions proposed by~\cite{Wolpert_HighLevel} are task-specific in the sense that they aim to predict an observation based on $Z_t$ from the aggregated process.}

\rev{
Closely related to Markov aggregation is the topic of \emph{lumpability}, i.e., the question whether a non-injective function of a Markov chain is Markov. Initial research in this area has performed by Kemeny and Snell (strong and weak lumpability,~\cite[\S6.3-6.4]{Kemeny_FMC}), Rosenblatt (lumpability of continuous-valued Markov processes~\cite{Rosenblatt_MarkovianFunctions}), and Buchholz (exact lumpability~\cite{Buchholz_Exact}). Gurvits and Ledoux discovered linear-algebraic conditions on the transition probability matrix of $\{Z_t{:}\ t = 1, 2,\dots\}$ and the aggregation function for weak and strong lumpability~\cite{GurvitsLedoux_MarkovPropertyLinearAlgebraApproach}. An equivalent characterization of strong lumpability in information-theoretic terms has been presented by Geiger and Temmel and Pfante et al.\  in~\cite{GeigerTemmel_kLump} and~\cite{Pfante_LevelID}, respectively. This information-theoretic characterization was used in a cost function for Markov aggregation in~\cite{GeigerEtAl_OptimalMarkovAggregation}.
}

\section{Generalized Information-Theoretic Markov Aggregation}\label{sec:MarkovAggregation}

Suppose $\{Z_t{:}\ t = 1, 2,\dots\}$ is a discrete-time, first-order, stationary Markov chain with finite alphabet $\mathcal{Z}$ and state transition matrix $\Amat=[A_{ij}]$, where 
\begin{equation}
\forall i,j \in \mathcal{Z}, t>1{:}\quad A_{ij} := \Pr (Z_t = j | Z_{t-1} = i ).
\end{equation}
Throughout this work we assume that $\Amat$ is irreducible. The Markov aggregation problem is concerned with finding a function $\zeta{:}\ \mathcal{Z} \rightarrow \overline{\mathcal{Z}}$, where typically $|\mathcal{Z}| \gg |\overline{\mathcal{Z}}|$, such that the reduced model captures \emph{relevant} aspects of the original model. Specifically, the authors of~\cite{Amjad_GeneralizedMA} suggest trading between two different objectives: The objective to make the process $\{\zeta(Z_t)\}$ as close to a Markov chain as possible, and the objective that $\{\zeta(Z_t)\}$ preserves the temporal dependence structure of the original Markov chain $\{Z_t\}$. They propose the following information-theoretic cost function for Markov aggregation:

\begin{definition}[Generalized Markov Aggregation~\cite{Amjad_GeneralizedMA}]\label{def:MA_I}
Let $\{Z_t\}$ be a discrete-time, stationary Markov chain with alphabet $\mathcal{Z}$ and state transition matrix $\Amat$, and suppose the set $\overline{\mathcal{Z}}$ is given. Let $\beta\in[0,1]$. The generalized information-theoretic Markov aggregation problem concerns finding a minimizer $\hat{\zeta}$ of
\begin{equation}
 \min_{\zeta{:}\ \mathcal{Z} \rightarrow \overline{\mathcal{Z}}} \mathcal{L}_{\beta} (\zeta)
\end{equation}
where the minimization is over all functions $\zeta{:}\ \mathcal{Z}\to\overline{\mathcal{Z}}$ and where, with $\overline{Z}_t:=\zeta(Z_t)$ for every $t\ge 1$,
\begin{multline}
	\mathcal{L}_{\beta}(\zeta) := \beta\mutinf{Z_1;Z_2} +(1-2\beta)\mutinf{Z_1;\overline{Z}_2} \\- (1-\beta)\mutinf{\overline{Z}_1;\overline{Z}_2}.
\end{multline} 
\end{definition}
For $\beta=1$, the cost function is reminiscent of the IB functional~\cite{Tishby_InformationBottleneck}, where compression is enforced by limiting the alphabet size of the compressed variable. For $\beta=0$, \rev{the cost function is linked to the phenomenon of lumpability} and $\zeta$ is chosen such that the process $\{\overline{Z}_t\}$ is ``as Markov as possible''; indeed, if $\mathcal{L}_{0} (\zeta)=0$, then $\{\overline{Z}_t\}$ is a Markov chain~\cite[Th.~1]{GeigerEtAl_OptimalMarkovAggregation}. Finally, it can be shown that minimizing $\mathcal{L}_{\frac{1}{2}} (\zeta)$ is equivalent to maximizing $I(\overline{Z}_1;\overline{Z}_2)$; essentially, this means that one wants to predict $\overline{Z}_2$ from $\overline{Z}_1$ with high accuracy, i.e., the temporal dependence structure should be preserved. This cost function was considered in~\cite{Meyn_MarkovAggregation} and was shown to be related to spectral clustering.

In the spirit of the IB formalism, mutual information can be used to measure relevance. Relevant information loss measures the information about some relevant RV $S$ that is lost by processing a statistically related RV $Z$ in a deterministic function $\zeta$. The quantity was introduced by Plumbley in the context of unsupervised neural networks~\cite{Plumbley_TN}:

\begin{definition}[Relevant Information Loss]\label{def:relloss}
Let $S$ and $Z$ be RVs with finite alphabet, and let $\zeta$ be a function defined on the alphabet $\mathcal{Z}$ of $Z$. Then, the relevant information loss w.r.t.\ $S$ that is induced by $\zeta$ is
\begin{equation}
\loss[S]{Z \rightarrow \zeta(Z)} := \mutinf{S; Z} - \mutinf{S; \zeta(Z)} = \mutinf{S;Z | \zeta(Z)} \geq 0. 
\end{equation}
\end{definition}

With this definition, we can rewrite the cost function for Markov aggregation in terms of relevant information loss:

\begin{lem}\label{lem:MA_RIL}
In the setting of Definition~\ref{def:MA_I} we have
\begin{equation}
\label{markovcost}
\mathcal{L}_{\beta}(\zeta) = \beta \loss[Z_1]{Z_2 \rightarrow \overline{Z}_2} + (1 - \beta) \loss[\overline{Z}_2]{Z_1 \rightarrow \overline{Z}_1}.
\end{equation}
\end{lem}

\rev{The function $\zeta$ partitions the alphabet $\dom{Z}$ into clusters. Hence, the first term captures how much information is lost about $Z_1$ if $Z_2$ is clustered via $\zeta$, while the second term captures how much information is lost about the \emph{cluster} $\overline{Z}_2$ if $Z_1$ is clustered via $\zeta$. This formulation will be our starting point for developing an information-theoretic cost function for co-clustering.}

\section{Information-Theoretic Co-Clustering via Markov Aggregation}\label{sec:CostFunctions}
We now turn to the co-clustering problem. Suppose we have two disjoint finite sets $\dom{X}$ and $\dom{Y}$ and a $|\dom{X}|\times|\dom{Y}|$ matrix $\Wvec$ containing, e.g., similarities, the number of co-occurrences, or correlations between elements of these two sets. As an example, if $\dom{X}$ is a set of documents and $\dom{Y}$ a set of words, then the $(i,j)$-th entry of $\Wvec$ could be the number of times the word $j$ appeared in document $i$. Co-clustering is concerned with finding partitions of $\dom{X}$ and $\dom{Y}$ (document and word clusters in this example), sacrificing information about the individual data elements to make the group characteristics more prominent and accessible.

\subsection{Adapting the Cost Function}

If the matrix $\Wvec$ is nonnegative, we can interpret it as the weight matrix of an undirected, weighted, bipartite graph, cf.~\cite{Dhillon_Graph}. Throughout this work we will assume that $\Wvec$ is such that the bipartite graph is irreducible. On this graph, one can then define a simple random walk, i.e., a Markov chain $\{Z_t\}$ with alphabet $\dom{X}\cup\dom{Y}$ and state transition matrix
\begin{equation}\label{eq:cocA}
\Amat = \mathbf{D}^{-1} \left[\begin{array}{cc}
            0 & \Wvec\\ \Wvec^T & 0
           \end{array}\right]
\end{equation}
where $\mathbf{D}$ is a diagonal matrix collecting sums of all connected edge weights of respective nodes. The matrix $\mathbf{D}$ normalizes each row of $\Amat$ to make it a probability distribution. Since the \rev{graph is} bipartite and undirected, the Markov chain $\{Z_t\}$ is \rev{2-periodic and} reversible. 

We now apply the Markov aggregation framework from Definition~\ref{def:MA_I} and Lemma~\ref{lem:MA_RIL} to the co-clustering problem. To this end, we add the constraint that the function $\zeta$ from Definition~\ref{def:MA_I} does not put elements of $\dom{X}$ and $\dom{Y}$ in the same cluster. This mutual exclusivity constraint guarantees that there exist functions $\Phi$ and $\Psi$ such that
\begin{equation}\label{eq:mutex}
 \forall i \in \dom{X}\cup\dom{Y}{:}\quad \zeta(i) = \begin{cases}
                                                              \Phi(i), & i\in\dom{X}\\
                                                              \Psi(i), & i\in\dom{Y}.
                                                             \end{cases}
\end{equation}

The following proposition transfers the cost function from Lemma~\ref{lem:MA_RIL} to the co-clustering setting:

\begin{prop}\label{prop:coc}
Suppose two disjoint finite sets $\dom{X}$ and $\dom{Y}$ and a nonnegative $|\dom{X}|\times|\dom{Y}|$ matrix $\Wvec$ containing similarities between elements of these two sets are given. Define two discrete RVs $X$ and $Y$ over these sets, where the joint distribution $P_{X,Y}$ is obtained by normalizing $\Wvec$. Let $\{Z_t\}$ be a stationary Markov chain with alphabet $\dom{X}\cup\dom{Y}$ and state transition matrix $\Amat$ given in~\eqref{eq:cocA}. Let $\beta\in[0,1]$ and suppose the sets $\overline{\dom{X}}$ and $\overline{\dom{Y}}$ are given. 

For every function $\zeta{:}\ \dom{X}\cup\dom{Y}\to \overline{\dom{X}}\cup\overline{\dom{Y}}$ satisfying the mutual exclusivity constraint~\eqref{eq:mutex}, we have
\begin{multline}
2\cdot\mathcal{L}_{\beta} (\zeta) = \beta (L_X(Y \rightarrow \overline{Y}) + L_Y(X \rightarrow \overline{X})) \\+ (1 - \beta) (L_{\overline{X}}(Y \rightarrow \overline{Y}) + L_{\overline{Y}}(X \rightarrow \overline{X})) =:\mathcal{L}_{\beta} (\Phi, \Psi)
\end{multline}
where $\overline{X}:=\Phi(X)$ and $\overline{{Y}}:=\Psi(Y)$.
\end{prop}

\begin{IEEEproof}
Suppose that $\{Z_t\}$ is a Markov chain with state space $\dom{X}\cup\dom{Y}$ and state transition matrix $\Amat$ as in~\eqref{eq:cocA}, with $\mathbf{D}$ given by
\begin{equation}
 \mathbf{D}=\mathrm{diag}\left(\left[\begin{array}{cc}
            0 & \Wvec\\ \Wvec^T & 0
           \end{array}\right] \mathbf{1}\right)
\end{equation}
where $\mathbf{1}$ is a vector of ones of appropriate length. Suppose $\boldsymbol{\mu}=[\mu_i]$ is the invariant distribution of $\Amat$, i.e., $\boldsymbol{\mu}^T=\boldsymbol{\mu}^T \Amat$. It follows that $\mathrm{diag}(\boldsymbol{\mu})\propto\mathbf{D}$. Suppose further that $P_{X,Y}$ is the joint distribution obtained by normalizing $\Wvec$. Then, the marginal distributions for $X$ and $Y$ are $P_X=\sum_{y\in\dom{Y}} P_{X,Y}(\cdot,y)\propto \Wvec\mathbf{1}$ and $P_Y^T=\sum_{x\in\dom{X}} P_{X,Y}(x,\cdot)\propto\mathbf{1}^T \Wvec$, respectively. \rev{From the 2-periodicity of $\{Z_t\}$ thus follows that}
\begin{equation}\label{eq:invariant}
 \mu_i =\frac{1}{2}\begin{cases}
     P_X(i), & i\in\dom{X}\\
     P_Y(i), & i\in\dom{Y}.
    \end{cases}
\end{equation}

Now assume that the Markov chain $\{Z_t\}$ is stationary, i.e., the distribution of $Z_1$ coincides with the invariant distribution $\boldsymbol{\mu}$. Let $U$ be a RV that indicates whether $Z_1$ was drawn from $\dom{X}$ or $\dom{Y}$, i.e.,
\begin{equation}
 U:=\begin{cases}
     1, & Z_1\in\dom{X}\\
     0, & Z_1\in\dom{Y}.
    \end{cases}
\end{equation}
\rev{Note that $U$ is a function not only of $Z_1$ but, by periodicity, of $Z_t$ for every $t$. The RV $U$ thus connects $P_{Z_t}$ with $P_X$ or $P_Y$; e.g., if $U=1$, then $P_{Z_3}=P_X$.}
It follows from~\eqref{eq:invariant} that $\Pr (U=1)=\Pr(U=0)=\frac{1}{2}$.

Finally, suppose that $\zeta$ satisfies the mutual exclusivity constraint~\eqref{eq:mutex}; hence $\Phi(\dom{X})=\overline{\dom{X}}$, $\Psi(\dom{Y})=\overline{\dom{Y}}$, and $U=1$ if and only if $\overline{Z}_1\in\overline{\dom{X}}$.

We now investigate $\mutinf{\tilde{Z}_1;\tilde{Z}_2}$, where $\tilde{Z}_i$ is either $Z_i$ or $\overline{Z}_i$. We get
\begin{align}
 &\mutinf{\tilde{Z}_1;\tilde{Z}_2} \stackrel{(a)}{=} \mutinf{\tilde{Z}_1,U;\tilde{Z}_2}\notag\\
 &\stackrel{(b)}{=} \mutinf{\tilde{Z}_1;\tilde{Z}_2|U} + \mutinf{U;\tilde{Z}_2}\notag\\
 &\stackrel{(c)}{=} \frac{1}{2} \mutinf{\tilde{Z}_1;\tilde{Z}_2|U=1}+\frac{1}{2} \mutinf{\tilde{Z}_1;\tilde{Z}_2|U=0} + \ent{U}
\end{align}
where $(a)$ is because $U$ is a function of $Z_1$ and $\overline{Z}_1$, $(b)$ is the chain rule of mutual information, and $(c)$ follows because $U$ is also a function of $Z_2$ and $\overline{Z}_2$ and from the definition of conditional mutual information. 

Now suppose $\tilde{Z}_1=\overline{Z}_1$ and $\tilde{Z}_2=Z_2$. If $U=1$, then $\overline{Z}_1\in\overline{\dom{X}}$ and $Z_2\in\dom{Y}$, and the joint distribution $P_{\overline{Z}_1,Z_2}$ equals the joint distribution $P_{\overline{X},Y}$. With similar considerations for $U=0$ we hence get
\begin{subequations}
 \begin{multline}
 \mutinf{\overline{Z}_1;Z_2} \\=  \frac{1}{2} \mutinf{\overline{Z}_1;{Z}_2|U=1}+\frac{1}{2} \mutinf{\overline{Z}_1;{Z}_2|U=0} + \ent{U}\\
 = \frac{1}{2} \mutinf{\overline{X};Y}+\frac{1}{2} \mutinf{X;\overline{Y}} + \ent{U}.
\end{multline}
Along the same lines we obtain
\begin{align}
  I({Z}_1;{Z}_2) &= I(X;Y)+\ent{U},\\
  I(\overline{Z}_1;\overline{Z}_2) &= I(\overline X;\overline Y)+\ent{U},\\
  I(Z_1;\overline{Z}_2) &= \frac{1}{2} I(X;\overline{Y})+ \frac{1}{2}I(\overline{X};Y) + \ent{U}.
\end{align}
\end{subequations}
Inserting these in the cost function in Lemma~\ref{lem:MA_RIL} and applying the definition of relevant information loss in Definition~\ref{def:relloss} completes the proof.
\end{IEEEproof}

We now present our cost function for information-theoretic co-clustering:

\begin{definition}[Generalized Information-Theoretic Co-Clustering]\label{def:coc}
The generalized information-theoretic co-clustering problem concerns finding a minimizer $( \hat{\Phi}, \hat{\Psi} )$ of
\begin{equation}\label{eq:gitcccost}
\min_{\Phi{:}\ \dom{X} \rightarrow \overline{\dom{X}},\ \Psi{:}\ \dom{Y} \rightarrow \overline{\dom{Y}}} \mathcal{L}_{\beta} (\Phi, \Psi)
\end{equation} 
where the minimization is over all functions $\Phi{:}\ \dom{X}\to \overline{\dom{X}}$ and $\Psi{:}\ \dom{Y}\to \overline{\dom{Y}}$ and where $\mathcal{L}_{\beta} (\Phi, \Psi)$ is as in the setting of Proposition~\ref{prop:coc}.
\end{definition}

The presented cost function admits an intuitive explanation for the effect of the parameter $\beta$: In the context of the words/documents co-clustering example above, minimizing $\loss[X]{Y \rightarrow \overline{Y}}$ means that we are looking for word clusters that tell us much about documents. In contrast, minimizing $\loss[\overline{X}]{Y \rightarrow \overline{Y}}$ means that we are looking for word and document clusters such that the word clusters tell us much about the document clusters. The parameter $\beta$ thus determines how strongly the two clusterings should be coupled. We show in Sections~\ref{sec:pathological}~\ref{sec:experiments} that the choice of $\beta$ can have a prominent effect on the clustering performance.

\subsection{Adapting a Sequential Optimization Heuristic}
\rev{In general,  finding a minimizer of our cost function~\eqref{eq:gitcccost} is a combinatorial problem with exponential computational complexity in $|\dom{X}|$ and $|\dom{Y}|$ . Hence heuristics for combinatorial or non-convex optimization are used to find good sub-optimal solutions with reasonable complexity}. In particular, it can be optimized by adapting heuristics proposed for information-theoretic co-clustering by other authors (see Sections~\ref{sec:relatedwork} and~\ref{sec:specialAlgos}). Since our cost function is derived from the generalized information-theoretic Markov aggregation problem, co-clustering solutions can be obtained by employing the aggregation algorithm proposed in~\cite{Amjad_GeneralizedMA} taking into account the additional mutual exclusivity constraint. The algorithm is a simple sequential heuristic for minimizing $\mathcal{L}_\beta$, similar to the sequential IB algorithm proposed in~\cite{Slonim_DocClustering} and the algorithm proposed by Dhillon et al.\ for information-theoretic co-clustering~\cite{Dhillon_ITClustering}. This algorithm is random in the sense that it is started with two random functions $\Phi$ and $\Psi$ with desired output cardinalities. In each iteration, these two functions are altered successively in order to reduce the cost function, either until we reach a maximum number of iterations or until the cost function has converged to within a chosen threshold of a local minimum. \rev{The authors of~\cite{Amjad_GeneralizedMA} introduced an annealing procedure for the $\beta$-parameter to escape local optima, which is particularly important for small values of $\beta$. The pseudocodes for the sequential heuristic, \textsc{sGITCC}, and the annealing heuristic, \textsc{AnnITCC}, are given in Algorithms~\ref{sGITCC} and~\ref{betaCC}, respectively; for details, the reader is referred to~\cite{Amjad_GeneralizedMA}. It can be shown along the lines of the corresponding result in~\cite{Amjad_GeneralizedMA} that, by storing intermediate results, the computational complexity of computing $\mathcal{L}_{\beta} (\Phi, \Psi_{\ell})$ and $\mathcal{L}_{\beta} (\Phi_j, \Psi)$ can be brought down to $\mathcal{O}(|\dom{X}|)$ and $\mathcal{O}(|\dom{Y}|)$, respectively. Thus, one iteration of Algorithm~\ref{sGITCC} has computational complexity of $\mathcal{O}(|\dom{X}| \cdot |\dom{Y}|\cdot \max\{|\overline{\dom{Y}}|,|\overline{\dom{X}}|\})$.}

%

\begin{algorithm}
\caption{Sequential Generalized Information-Theoretic Co-Clustering (\textsc{sGITCC})}\label{sGITCC}
\begin{algorithmic}[1]
\Function {$(\Phi, \Psi) = $ sGITCC}{$P_{X,Y}$, $\beta$, $|\overline{\dom{X}}|$, $|\overline{\dom{Y}}|$, $\numitermax$, $\toler$, optional: initial clustering $(\Phi_{\text{init}}, \Psi_{\text{init}})$}
\If {$(\Phi_{\text{init}}, \Psi_{\text{init}})$ is empty} \Comment \textit{Inizialization}
\State $(\Phi, \Psi) \gets $ Random Clustering
\Else
\State $(\Phi, \Psi) \gets (\Phi_{\text{init}}, \Psi_{\text{init}})$
\EndIf
\State $\numiter \gets 0$
\While {$\numiter < \numitermax \land \delta > \toler$} \Comment \textit{Main Loop}
\State $C_{old} \gets \mathcal{L}_{\beta} (\Phi, \Psi)$
\For {all elements $i \in \dom{X}$} \Comment \textit{Optimizing $\Phi$}
    \For {all clusters $j \in \overline{\dom{X}}$}
       \State $\Phi_j(x) = 
       \begin{cases}
       \Phi(x)  &\quad \forall x \neq i \\
       j        &\quad x=i		                                          
       \end{cases}$                             
    \EndFor
    \State $\Phi(i) = \argmin\limits_{j} \mathcal{L}_{\beta} (\Phi_j, \Psi)$
\EndFor
\For {all elements $k \in \dom{Y}$} \Comment \textit{Optimizing $\Psi$}
    \For {all clusters $\ell \in \overline{\dom{Y}}$}
       \State $\Psi_{\ell}(y) = 
       \begin{cases}
       \Psi(y)  &\quad \forall y \neq k \\
       \ell        &\quad y=k		                                          
       \end{cases}$                             
    \EndFor
    \State $\Psi(k) = \argmin\limits_{\ell} \mathcal{L}_{\beta} (\Phi, \Psi_{\ell})$
\Comment \textit{Break ties}
\EndFor
\State $\delta \gets C_{old} - \mathcal{L}_{\beta} (\Phi, \Psi)$
\State $\numiter \gets \numiter + 1$
\EndWhile
\EndFunction
\end{algorithmic}
\end{algorithm}

\begin{algorithm}
\caption{$\beta$-Annealing Information-Theoretic Co-Clustering (\textsc{AnnITCC})}\label{betaCC}
\begin{algorithmic}[1]
\Function {$(\Phi, \Psi) =$ AnnITCC}{$P_{X,Y}$, $\beta$, $|\overline{\dom{X}}|$, $|\overline{\dom{Y}}|$, $\numitermax$, $\toler$, $\Delta$}
\State $\alpha \gets 1$
\State $(\Phi, \Psi)$ = sGITCC($P_{XY}$, $\beta$, $|\overline{\dom{X}}|$,$|\overline{\dom{Y}}|$, $\numitermax$, $\toler$) 
\While {$\alpha > \beta$}
\State $\alpha \gets \max\{\alpha - \Delta,\beta\}$ 
\State $(\Phi, \Psi)$ = sGITCC($P_{XY}$, $\alpha$, $|\overline{\dom{X}}|$, $|\overline{\dom{Y}}|$, $\numitermax$, $\toler$, $(\Phi, \Psi)$)
\EndWhile
\EndFunction
\end{algorithmic}
\end{algorithm}

The following example shows how the sequential heuristic in Algorithm~\ref{sGITCC} can get stuck in a poor local optimum for $\beta = \frac{1}{2}$. The same example is unproblematic for $\beta=1$. Since one can certainly find heuristics that perform optimally in this example even for $\beta=\frac{1}{2}$, matching the heuristic to the cost function seems to be an important issue. We will see further evidence for the impact of heuristics on performance in our experiments with the Newsgroup20 dataset in Section~\ref{sec:NG20}.
\begin{example}
 Consider the following $3\times 4$ matrix describing the joint probability distribution between $X$ and $Y$:
\begin{align*}
P_{X,Y} &= \left[\begin{array}{c|c"cc}
0.25 & 0 & 0& 0 \\
\hline
0 & 0.25 & 0 & 0 \\
\thickhline
0 & 0 & 0.25 & 0.25
\end{array}\right]
\end{align*}
We are interested in two row clusters and two column clusters, i.e., $|\overline{\dom{X}}|=|\overline{\dom{Y}}|=2$. Suppose that during some iteration, the clustering functions $\Phi$ and $\Psi$ induce the partition indicated by the thin black lines in the matrix $P_{X,Y}$. At this stage, for $\beta=\frac{1}{2}$ the sequential algorithm will terminate since this $\Phi$ is the optimal choice for $\Psi$ fixed, and this $\Psi$ is the optimal choice for $\Phi$ fixed. In other words, changing either clustering function alone increases the cost $\mathcal{L}_{\frac{1}{2}}=I(X;Y)-I(\overline{X}; \overline{Y})$. Nevertheless, it is clear from looking at $P_{X,Y}$, that the cost is minimized ($I(\overline{X}; \overline{Y})$ is maximized) for the partition indicated by the thick black lines. The algorithm thus gets stuck for $\beta = \frac{1}{2}$ because the cost function in this case only depends on the clustered variables, and because it updates the clustering functions subsequently rather than jointly. For larger values of $\beta$, the coupling between the clustering functions is weaker. In particular, for $\beta=1$, the clustering functions can be optimized independently of each other, and the algorithm hence terminates at a partition consistent with the vertical thick line, even if it was started at the partition indicated by the thin lines.
\end{example}

\section{Special Cases of Generalized Information-Theoretic Co-Clustering}\label{sec:specialAlgos}

We next show that our generalized information-theoretic co-clustering cost function from Definition~\ref{def:coc} contains, for appropriate settings of the parameter $\beta$, previously proposed cost functions as special cases. For example, for $\beta=1$, we obtain
\begin{equation}
\mathcal{L}_{1} (\Phi, \Psi) = L_X(Y \rightarrow \overline{Y}) + L_Y (X \rightarrow \overline{X}).
\end{equation}
This cost function consists of two IB functionals: The first term considers clustering $Y$ with $X$ the relevant variable, while the second term considers clustering $X$ with $Y$ the relevant variable. This approach rewards clustering solutions for $X$ and $Y$ that are completely decoupled. To minimize this cost function, one can use the fixed-point equations derived in~\cite{Tishby_InformationBottleneck} or the agglomerative IB method (aIB) that merges clusters until the desired cardinality is reached~\cite{Slonim_Agglomerative}. Finally, a sequential IB method (sIB) has been proposed that iteratively moves an element from its current cluster to the cluster that minimizes the cost until a local minimum is reached~\cite{Slonim_DocClustering}.

More interestingly, we can rewrite the cost function that Dhillon et al.\ proposed in~\cite{Dhillon_ITClustering} for information-theoretic co-clustering (ITCC) and obtain
\begin{equation}\label{eq:compITCC}
 \mathcal{L}_{\text{ITCC}} (\Phi, \Psi) := I(X;Y) - I(\overline{X}; \overline{Y})= \mathcal{L}_{\frac{1}{2}} (\Phi, \Psi).
\end{equation}
Thus, ITCC is a special case of our cost function for $\beta=\frac{1}{2}$. The authors of~\cite{Dhillon_ITClustering} proposed a sequential algorithm, similar to sIB, alternating between optimizing $\Phi$ and $\Psi$. \rev{Furthermore, $\mathcal{L}_{\text{ITCC}} (\Phi, \Psi)$ can be optimized via non-negative matrix tri-factorization~\cite[Lemma~2.1]{Dhillon_ITClustering} and thus yields a generative model as a result. We are not aware if a similar connection to generative models holds for other values of $\beta$.}

In~\cite{Bekkerman_MultiClustering}, the cost function $\mathcal{L}_{\frac{1}{2}}$ is generalized to pairwise interactions of multiple variables (the two-dimensional case is equivalent to co-clustering). The authors introduce a multilevel heuristic that schedules the splitting of clusters, merges clusters following the ideas of aIB~\cite{Slonim_DoubleClustering}, and optimizes intermediate results sequentially with sIB.

The authors of \cite{Slonim_DoubleClustering} proposed applying aIB twice to obtain the co-clustering. In the first step, in which the set $\dom{X}$ is clustered, they treat $Y$ as the relevant variable; in the second step, in which the set $\dom{Y}$ is clustered, they treat the clustered variable $\overline{X}$ as relevant. In essence, the authors of \cite{Slonim_DoubleClustering} thus minimize the functional
\begin{equation}\label{eq:aibdouble}
\mathcal{L}_{\text{IB-double}}(\Phi, \Psi) = L_Y (X \rightarrow \overline{X}) + L_{\overline{X}}(Y \rightarrow \overline{Y}) = \mathcal{L}_{\frac{1}{2}} (\Phi, \Psi)
\end{equation}
in a greedy manner: They first optimize over $\Phi$ to minimize only the first term and then optimize over $\Psi$ to minimize the second term. Comparing~\eqref{eq:compITCC} and~\eqref{eq:aibdouble} reveals that~\cite{Slonim_DoubleClustering} and~\cite{Dhillon_ITClustering} optimize the same cost function; the fact that they report different performance results can only be explained by differences in the optimization heuristic and (possibly) preprocessing steps. We will elaborate on this topic in our experiments with the Newsgroup20 dataset in Section~\ref{sec:NG20}.

Another approach related to IB, called information bottleneck co-clustering (IBCC), was proposed in~\cite{Wang_IBCC}. The functional being maximized by IBCC is
\begin{multline}
\label{community338}
\mathcal{L}_{\text{IBCC}}(\Phi, \Psi)  := I(X; \overline{Y}) + I(\overline{X}; Y) + I(\overline{X}; \overline{Y})\\
= 3I(X;Y)-2 \mathcal{L}_{\frac{3}{4}} (\Phi, \Psi).
\end{multline}
Hence, also IBCC is a special case of the generalized Markov aggregation framework for $\beta = \frac{3}{4}$. The authors of~\cite{Wang_IBCC} propose two algorithms: One is an agglomerative, i.e., a greedy merging algorithm, the other is an iterative update of fixed-point equations in the spirit of~\cite{Tishby_InformationBottleneck}.

Finally, for $\beta = 0$ we obtain the functional 
\begin{equation}
\mathcal{L}_{0} (\Phi, \Psi) = L_{\overline{X}}(Y \rightarrow \overline{Y}) + L_{\overline{Y}} (X \rightarrow \overline{X}).
\end{equation}
\rev{As previously mentioned, for Markov aggregation and $\beta=0$ the cost function is linked to the phenomenon of lumpability. In the co-clustering framework, lumpability means that the two clustering solutions that are coupled.} Precisely, we have $\mathcal{L}_{0} (\Phi, \Psi) =0$ if the rows $X$ and columns $Y$ do not share more information with the column clusters $\overline{Y}$ and row clusters $\overline{X}$, respectively, than the row clusters and column clusters share with each other. Unfortunately, we also have $\mathcal{L}_{0} (\Phi, \Psi) =0$ if $\overline{X}$ and $\overline{Y}$ are independent, which suggests an inherent drawback of $\mathcal{L}_0$ for co-clustering (despite its justification in Markov aggregation~\cite{GeigerEtAl_OptimalMarkovAggregation}). This leads to $\mathcal{L}_{0}$ (and, in general, $\mathcal{L}_\beta$ for small $\beta$) having multiple bad local optima in which any heuristic tends to get stuck.

\section{Strengths and Limitations of Generalized Information-Theoretic Co-Clustering}\label{sec:pathological}

\rev{In this section we use examples and experiments on synthetic datasets to highlight different aspects of using $\mathcal{L}_\beta$ and our proposed optimization heuristic for co-clustering. Specifically, we will point at limitations and strengths of co-clustering in comparison with one-sided clustering ($\beta=1$), which leads to guiding principles for the choice of $\beta$ depending on characteristics of the considered dataset.}

\subsection{Examples}
In the previous section we have discovered an inherent shortcoming of $\mathcal{L}_0$ in that it leads to co-clusterings with (near-)independent cluster RVs. In this subsection, we point at further limitations of information-theoretic cost functions for co-clustering. These shortcomings are independent of the employed optimization heuristic, but rather reflect that in some scenarios not even the global optimum of the cost function coincides with the ground truth (or an otherwise desired co-clustering solution). Sometimes this is simply caused by the fact that the cost function does not fit the underlying model -- e.g., if $\Wvec$ is generated according to a Poisson latent block model, then maximizing the likelihood of the co-clustering is equivalent to minimizing $\mathcal{L}_{\frac{1}{2}}$ only if the clusters have all the same cardinality~\cite[Sec.~2.2]{PLBM_Nadif}. In contrast, the following two scenarios make no assumptions on an underlying model but illustrate shortcomings inherent to the considered information-theoretic cost functions.

\subsubsection{Largely Different $|\overline{\dom{X}}|$ and $|\overline{\dom{Y}}|$}\label{sec:differentCardinalities}
An advantage of information-theoretic co-clustering approaches over, e.g., spectral~\cite{Dhillon_Graph,Ailem_Modularity} or certain block model-based approaches~\cite{PLBM_Nadif} is that the former admit different cardinalities for the clustered sets $|\overline{\dom{X}}|$ and $|\overline{\dom{Y}}|$. If, however, these cardinalities differ greatly, then minimizing $\mathcal{L}_\beta$ becomes problematic especially for small values of $\beta$. Let us assume w.l.o.g.\ that $|\overline{\dom{Y}}| < |\overline{\dom{X}}|$. Then, the optimization term $L_{\overline{Y}}({X\to\overline{X}})$ is limited by the information contained in $\overline{Y}$ rather than by the information loss induced by clustering $X$ to $\overline{X}$; many functions $\Phi$ may bring $L_{\overline{Y}}({X\to\overline{X}})$ close to zero simply because $\overline{Y}$ itself already contains little information. Similarly, the term $L_{\overline{X}}({Y\to\overline{Y}})$ may be large for many choices of $\Phi$, because, again, the limiting factor is the coarse clustering from $Y$ to $\overline{Y}$. These terms get more importance in~\eqref{eq:gitcccost} if $\beta$ is small. In other words, coupled co-clustering fails because the clustered variables contain little information. We illustrate this with a particular example, in which the joint probability distribution between $X$ and $Y$ is
\begin{align*}
P_{X,Y} &= \left[\begin{array}{cc"cc}
0.125 & 0 & 0& 0 \\
\hline
0.125 & 0 & 0& 0 \\
\thickhline
0 & 0.125 & 0 & 0 \\
0 & 0.125 & 0 & 0 \\
\thickhline
\hline
0 & 0 & 0.125 & 0 \\
0 & 0 & 0.125 & 0 \\
\thickhline
0 & 0 & 0 & 0.125\\
\hline
0 & 0 & 0 & 0.125
\end{array}\right].
\end{align*}
Our aim is to obtain a co-clustering with $|\overline{\dom{Y}}| =2$ and $|\overline{\dom{X}}| =4$. In $P_{X,Y}$, the thick vertical line indicates one possibility for $\Psi$ (a plausible ground truth). The horizontal lines indicate two possible options, $\Phi_1$ (thick lines) and $\Phi_2$ (thin lines) for the row clustering, where $\Phi_1$ corresponds to a plausible ground truth. 

For $\beta =1$, $(\Phi_1,\Psi)$ has a lower cost than $(\Phi_2,\Psi)$, as desired. Furthermore, one can show that $(\Phi_1,\Psi)$ minimizes the cost function; $\mathcal{L}_1$ has its global minimum at the ground truth.
For $\beta =\frac{1}{2}$, by evaluating $I(\overline{X}; \overline{Y})$ we see that both $(\Phi_1,\Psi)$ and $(\Phi_2,\Psi)$ have the same cost. In fact, any row clustering function $\Phi$ that shares the cluster boundary with the thick horizontal line in the middle has the same $I(\overline{X}; \overline{Y})$ for the given column clustering function $\Psi$: In this case, $\overline{X}$ determines $\overline{Y}$, hence we achieve the maximum $I(\overline{X}; \overline{Y})=H(\overline{Y})=1$; the cost function has multiple global minima, only one of which lies at the ground truth.
Finally, for $\beta=0$, $(\Phi_1,\Psi)$ has a higher cost than $(\Phi_2,\Psi)$. This implies that even if we initialize our algorithm at the ground truth (this could be the case if we do $\beta$-annealing) we move away from this clustering solution when we optimize the cost function for smaller values of $\beta$.

\subsubsection{Trading Entropy for Conditional Entropy}\label{sec:entVScond}

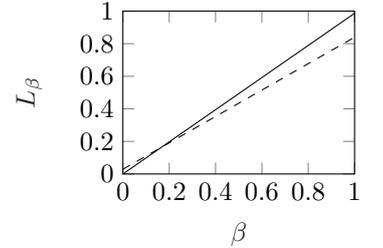
\begin{figure}
 \centering 
   \subfigure[\label{fig:trade:P1}]{\raisebox{2cm}{$\left[\begin{array}{c|c:c}
0.12 &0&    0\\
\hline
 0  &  0.39 &0.05\\
\hdashline
0   & 0.05 &0.39
\end{array}\right]$}}
 \hfill
   \subfigure[\label{fig:trade:cost1}]{
%
%
\begin{tikzpicture}

\begin{axis}[%
width=1.2in,
height=0.85in,
at={(0.758in,0.481in)},
scale only axis,
xmin=0,
xmax=1,
xlabel style={font=\color{white!15!black}},
xlabel={$\beta$},
ymin=0,
ymax=1,
ylabel style={font=\color{white!15!black}},
ylabel={$L_\beta$},
axis background/.style={fill=white}
]
\addplot [color=black, forget plot]
  table[row sep=crcr]{%
0	0\\
0.1	0.0861013431600937\\
0.2	0.172202686320187\\
0.3	0.258304029480281\\
0.4	0.344405372640375\\
0.5	0.430506715800468\\
0.6	0.516608058960562\\
0.7	0.602709402120656\\
0.8	0.688810745280749\\
0.9	0.774912088440843\\
1	0.861013431600937\\
};
\addplot [color=black, dashed, forget plot]
  table[row sep=crcr]{%
0	0.0366775702053233\\
0.1	0.116964481997853\\
0.2	0.197251393790382\\
0.3	0.277538305582912\\
0.4	0.357825217375441\\
0.5	0.438112129167971\\
0.6	0.5183990409605\\
0.7	0.59868595275303\\
0.8	0.678972864545559\\
0.9	0.759259776338088\\
1	0.839546688130618\\
};
\end{axis}
\end{tikzpicture}%
} 
   \subfigure[\label{fig:trade:P2}]{\raisebox{2cm}{$\left[\begin{array}{c|c:c}
0.12 &0&    0\\
\hline
 0  &  0.4 &0.04\\
\hdashline
0   & 0.04 &0.4
\end{array}\right]$}} \hfill
   \subfigure[\label{fig:trade:cost2}]{
%
%
\begin{tikzpicture}

\begin{axis}[%
width=1.2in,
height=0.85in,
at={(0.758in,0.481in)},
scale only axis,
xmin=0,
xmax=1,
xlabel style={font=\color{white!15!black}},
xlabel={$\beta$},
ymin=0,
ymax=1,
ylabel style={font=\color{white!15!black}},
ylabel={$L_\beta$},
axis background/.style={fill=white}
]
\addplot [color=black, forget plot]
  table[row sep=crcr]{%
0	0\\
0.1	0.0986485303018136\\
0.2	0.197297060603627\\
0.3	0.295945590905441\\
0.4	0.394594121207254\\
0.5	0.493242651509068\\
0.6	0.591891181810882\\
0.7	0.690539712112695\\
0.8	0.789188242414509\\
0.9	0.887836772716322\\
1	0.986485303018136\\
};
\addplot [color=black, dashed, forget plot]
  table[row sep=crcr]{%
0	0.029022857346849\\
0.1	0.110075240425226\\
0.2	0.191127623503603\\
0.3	0.272180006581979\\
0.4	0.353232389660356\\
0.5	0.434284772738733\\
0.6	0.51533715581711\\
0.7	0.596389538895487\\
0.8	0.677441921973864\\
0.9	0.758494305052241\\
1	0.839546688130617\\
};
\end{axis}
\end{tikzpicture}%
} 
\caption{Trading entropy for conditional entropy. (a) and (c) show joint distributions $P_{X,Y}$ together with two possible co-clusterings, while (b) and (d) show the corresponding values of the cost function for different values of $\beta$. Solid and dashed curves in (b) and (d) correspond to co-clusterings indicated by dashed and solid lines in (a) and (c).}
\end{figure}

\rev{Consider the joint distribution in Fig.~\ref{fig:trade:P1} that describes a dataset with a well-separated co-cluster structure for $|\overline{\dom{X}}|=|\overline{\dom{Y}}|=2$ (based on zeros and indicated by solid lines, denoted by $(\Phi^\bullet, \Psi^\bullet)$). We evaluate our cost function for different values of $\beta$, both for $(\Phi^\bullet, \Psi^\bullet)$ and for an alternative co-clustering indicated by dashed lines, denoted by $(\Phi, \Psi)$. It can be seen in Fig.~\ref{fig:trade:cost1} that, for $\beta \in [0.65,1]$, we have $\mathcal{L}_{\beta} (\Phi^\bullet, \Psi^\bullet) > \mathcal{L}_{\beta} (\Phi, \Psi)$, i.e., the ``incorrect'' solution has a lower cost than the ground truth. While in this case, e.g., ITCC~\cite{Dhillon_ITClustering} would probably terminate with $(\Phi^\bullet, \Psi^\bullet)$, it is easy to construct an example where ITCC fails. Changing our example only slightly leads to generalized information-theoretic co-clustering preferring  $(\Phi, \Psi)$ over $(\Phi^\bullet, \Psi^\bullet)$ for all $\beta$ in $[0.15, 1]$ (see Figs.~\ref{fig:trade:P2} and~\ref{fig:trade:cost2}).}

\rev{These examples show that even for datasets with a well-separated co-cluster structure, for a range of $\beta$ there can be (local and global) minima having a lower cost $\mathcal{L}_{\beta}$ than the ground truth. This can be explained by the fact that optimizing the cost function for a given value of $\beta$ boils down to maximizing/minimizing a combination of several mutual information terms. For example, for $\beta=\frac{1}{2}$ we aim to maximize, cf.~\eqref{eq:compITCC}
\begin{equation}
I(\overline{X}; \overline{Y})=H(\overline{X}) - H(\overline{X}|\overline{Y}).
\end{equation}
This leads to two competing goals: entropy maximization (preferring clusters with roughly equal probabilities) and conditional entropy minimization (preferring row clusters that determine column clusters, and vice-versa). For the range of $\beta$ where $\mathcal{L}_{\beta} (\Phi^\bullet, \Psi^\bullet)$ is not the global minimum, the first goal outweighs the second.} 

\rev{Note that for joint distributions with a well-separated co-cluster structure we have $\mathcal{L}_{0} (\Phi^\bullet, \Psi^\bullet) =0$ since $I(X; \overline{Y}) =I(\overline{X}; Y) = I(\overline{X}; \overline{Y})$. Nevertheless, due to the shortcoming discussed in Section~\ref{sec:specialAlgos}, this global optimum may not found because many other co-clusterings lead to $\mathcal{L}_{0} (\Phi, \Psi)\approx 0$. }

\subsection{Synthetic Datasets}\label{sec:synthetic}
\rev{Next, we perform experiments with two different synthetic datasets to explore further the relation between suitable choices of $\beta$ and the characteristics of the dataset.} Since our focus is on providing a better understanding of information-theoretic co-clustering, we assume that the true numbers of clusters and the true clustering functions $\Phi^\bullet$ and $\Psi^\bullet$ are known. As an accuracy measure, we employ the micro-averaged precision, which we define as follows:
\begin{equation}\label{eq:map}
\mathrm{MAP}(\Phi,\Phi^\bullet) := \max_{\pi} \frac{\sum_{j\in\overline{\dom{X}}} |\Phi^{-1}(j)\cap \Phi^{\bullet-1}(\pi(j)) | }{|\dom{X}|}
\end{equation}
where the maximization is over all permutations $\pi$ of the set $\overline{\dom{X}}$. The micro-averaged precision $\mathrm{MAP}(\Psi,\Psi^\bullet)$ is computed along the same lines. Note that $\mathrm{MAP}(\cdot,\cdot)$ requires that the clustering solution found by the algorithm has the same number of clusters as are present in the ground truth. Since we assume the true number of clusters to be known, this is unproblematic. If the number of clusters is unknown, one can resort to more sophisticated measures such as the adjusted Rand index or normalized mutual information. In the present case, all of these measures will lead to similar qualitative results.

\rev{Unless noted otherwise, we set $\toler=0$, $\numitermax=20$, and $\Delta=0.1$ and ran \textsc{AnnITCC} for values of $\beta$ between 0 and 1 in steps of 0.1. The simulation code for these and the real-world experiments in Section~\ref{sec:experiments} is publicly accessible.\footnote{bitbucket.org/bernhard\_geiger/coclustering\_markovaggregation}}

The first experiment looks at the clustering performance in the presence of noise. We generated a joint probability distribution $T_{X,Y}$ with 80 rows and 50 columns, i.e., $|\dom{X}|=80$ and $|\dom{Y}|=50$, and planted co-clusters such that $T_{X,Y}$ is constant within each co-cluster. A colorplot of $T_{X,Y}$ is shown in Fig.~\ref{fig:arti3:noisefree}. The figure also shows the ground truth $\Phi^\bullet$ ($|\overline{\dom{X}}|=5$) and $\Psi^\bullet$ ($|\overline{\dom{Y}}|=3$). We moreover constructed
a random probability distribution $N$ and constructed $P_{X,Y}$ from a weighted average of $T_{X,Y}$ and $N$, i.e., 
\begin{equation}
P_{X,Y} = (1 - \varepsilon) T_{X,Y} + \varepsilon N
\end{equation}
where $\varepsilon\in\{0,0.5,0.7,0.8\}$. Colorplots of $P_{X,Y}$ are shown in Fig.~\ref{fig:arti3:noise4} and~\ref{fig:arti3:noise8} for $\varepsilon=0.5$ and $\varepsilon=0.8$, respectively.

\newcommand{\rulesep}{\unskip\ \vrule\ }
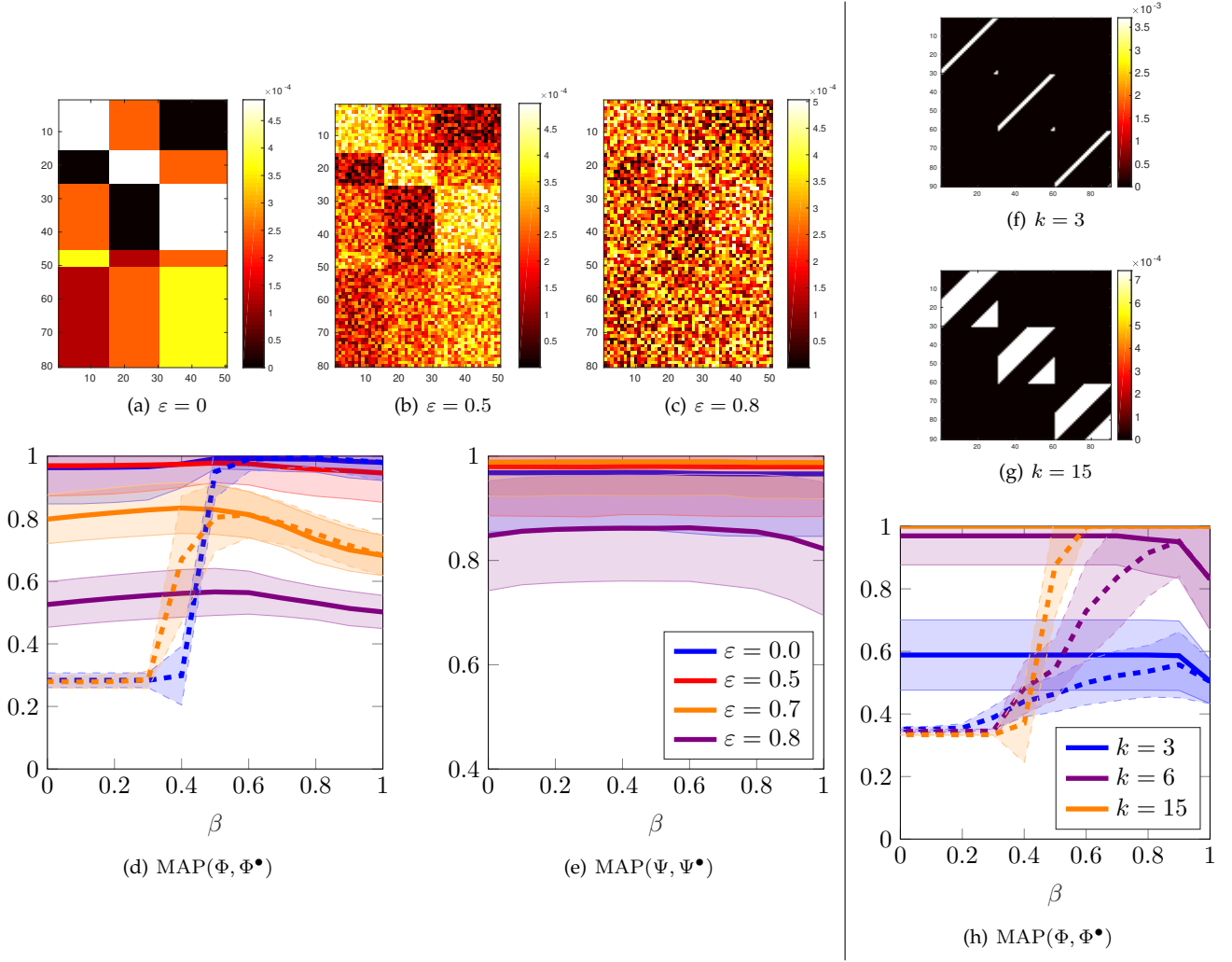
\begin{figure*}[t]
\centering 
	\begin{minipage}{0.65\textwidth}
	\hspace{0.5cm}
	\subfigure[$\varepsilon = 0$]{\includegraphics[width=0.3\textwidth]{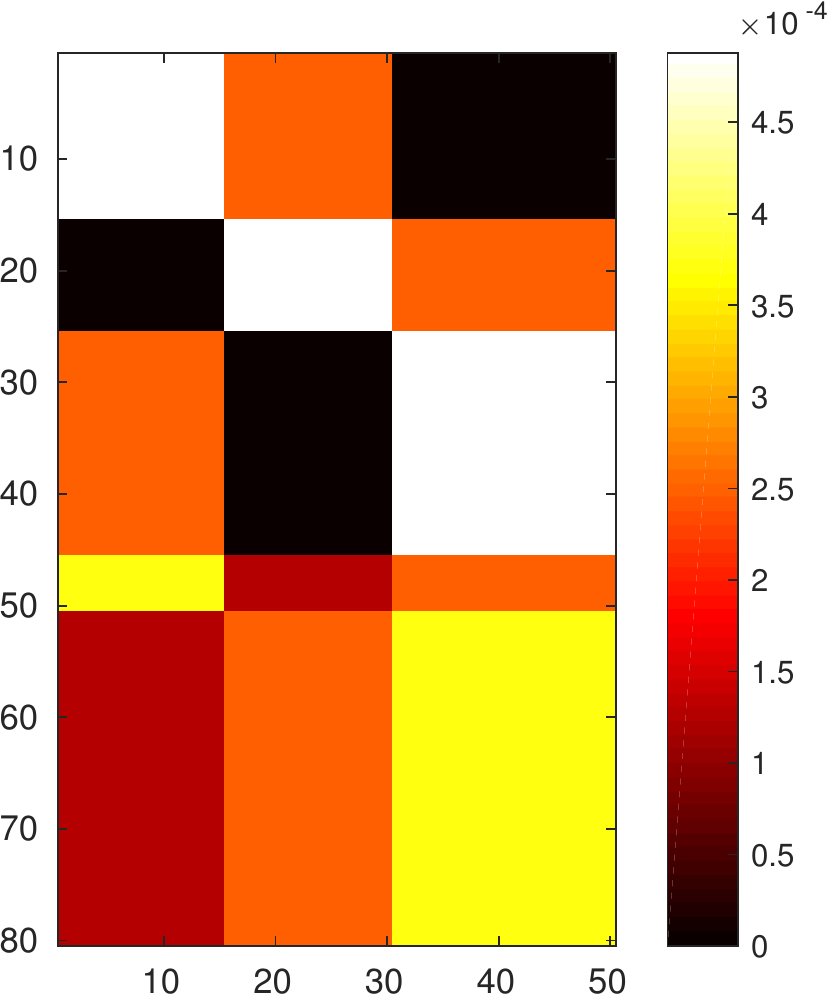}\label{fig:arti3:noisefree}} \hfill
    \subfigure[$\varepsilon = 0.5$]{\includegraphics[width=0.3\textwidth]{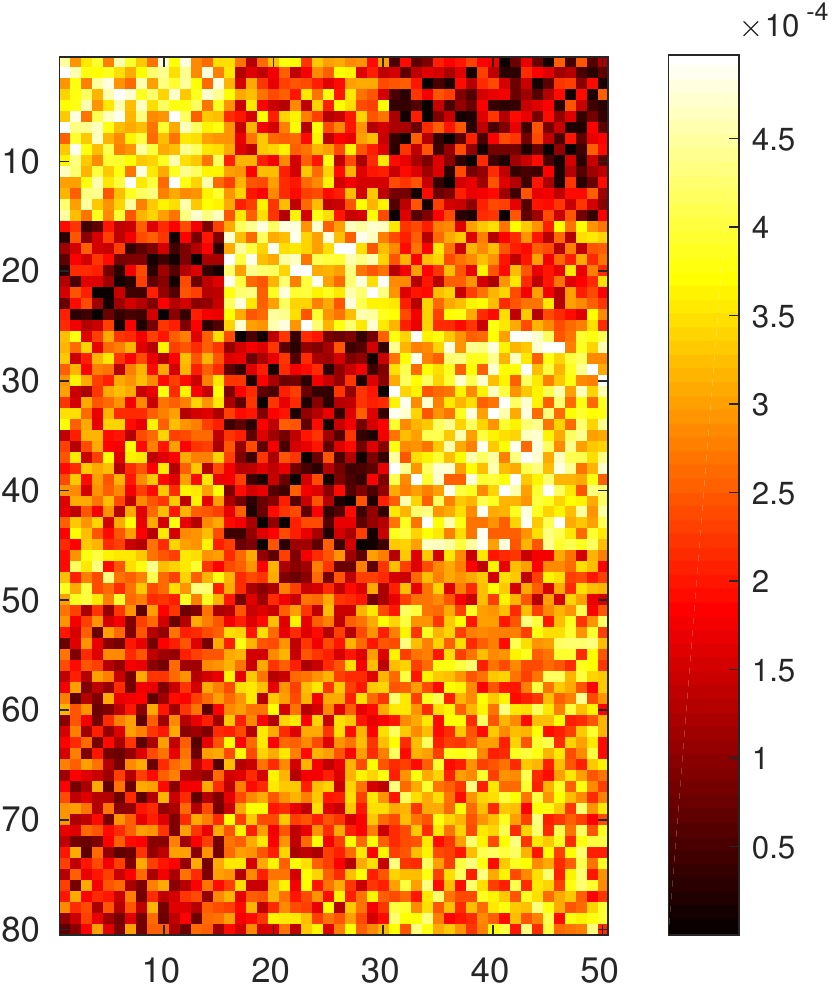}\label{fig:arti3:noise4}}\hfill
    \subfigure[$\varepsilon = 0.8$]{\includegraphics[width=0.3\textwidth]{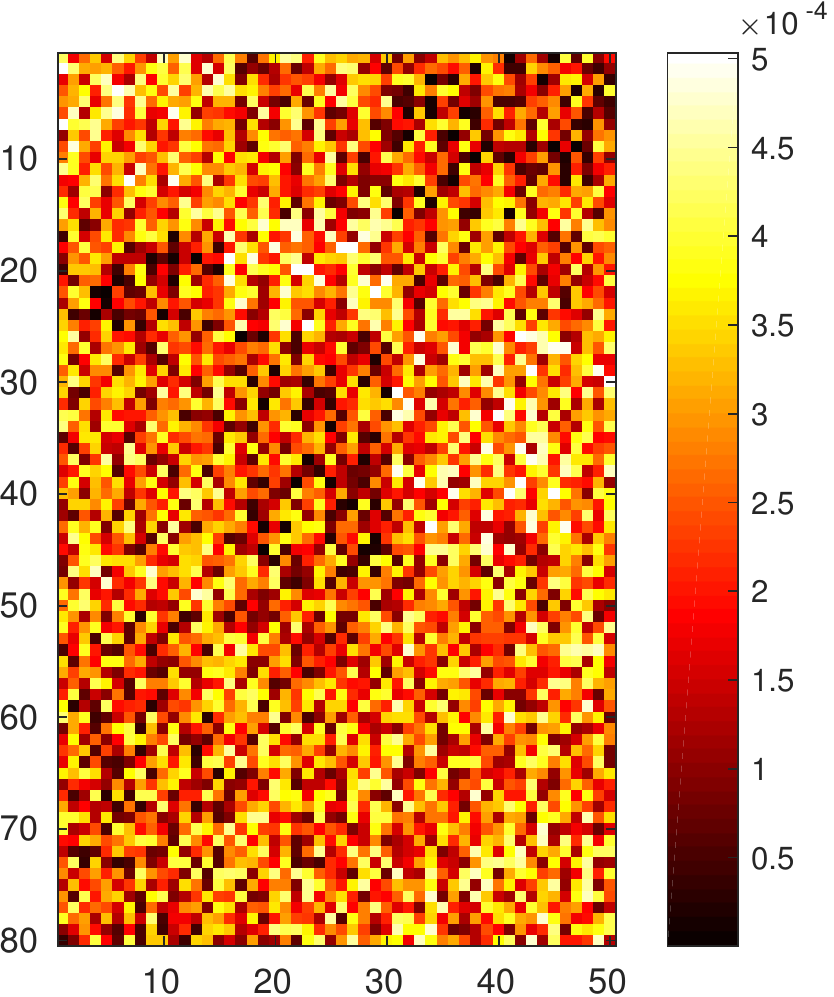}\label{fig:arti3:noise8}}\hspace{0.5cm}\\
    \subfigure[$ \mathrm{MAP}(\Phi,\Phi^\bullet)$]{
%
%

\begin{tikzpicture}

\begin{axis}[%
width=0.4\textwidth,
height=1.75in,
at={(0.758in,0.481in)},
scale only axis,
xmin=0,
xmax=1,
xlabel style={font=\color{white!15!black}},
xlabel={$\beta$},
ymin=0.0,
ymax=1,
axis background/.style={fill=white},
legend style={at={(0.97,0.03)}, anchor=south east, legend cell align=left, align=left, draw=white!15!black}
]

\addplot[area legend, draw=none, fill=blue, fill opacity=0.15, forget plot]
table[row sep=crcr] {%
x	y\\
1	0.921976440414829\\
0.9	0.930900832611193\\
0.8	0.948826465835472\\
0.7	0.951380492318666\\
0.6	0.956695732852549\\
0.5	0.959185624560606\\
0.4	0.89917602819783\\
0.3	0.860124843258398\\
0.2	0.852070681692164\\
0.1	0.847814401810966\\
0	0.847566368124498\\
0	1.0790336318755\\
0.1	1.07888559818903\\
0.2	1.07672931830783\\
0.3	1.0722751567416\\
0.4	1.05122397180217\\
0.5	1.02451437543939\\
0.6	1.02875426714745\\
0.7	1.03071950768133\\
0.8	1.03152353416453\\
0.9	1.03654916738881\\
1	1.03802355958517\\
}--cycle;
\addplot [color=white!55!blue, forget plot]
  table[row sep=crcr]{%
1	0.921976440414829\\
0.9	0.930900832611193\\
0.8	0.948826465835472\\
0.7	0.951380492318666\\
0.6	0.956695732852549\\
0.5	0.959185624560606\\
0.4	0.89917602819783\\
0.3	0.860124843258398\\
0.2	0.852070681692164\\
0.1	0.847814401810966\\
0	0.847566368124498\\
};
\addplot [color=white!55!blue, forget plot]
  table[row sep=crcr]{%
1	1.03802355958517\\
0.9	1.03654916738881\\
0.8	1.03152353416453\\
0.7	1.03071950768133\\
0.6	1.02875426714745\\
0.5	1.02451437543939\\
0.4	1.05122397180217\\
0.3	1.0722751567416\\
0.2	1.07672931830783\\
0.1	1.07888559818903\\
0	1.0790336318755\\
};
\addplot [color=blue, line width=2.0pt]
  table[row sep=crcr]{%
1	0.979999999999999\\
0.9	0.983724999999999\\
0.8	0.990175\\
0.7	0.99105\\
0.6	0.992725\\
0.5	0.99185\\
0.4	0.9752\\
0.3	0.9662\\
0.2	0.964399999999999\\
0.1	0.963349999999999\\
0	0.963299999999999\\
};

\addplot[area legend, draw=none, fill=red, fill opacity=0.15, forget plot]
table[row sep=crcr] {%
x	y\\
1	0.852625679250985\\
0.9	0.863700880215436\\
0.8	0.872136036426909\\
0.7	0.886561747691671\\
0.6	0.908070113047607\\
0.5	0.915823838337564\\
0.4	0.900536772751223\\
0.3	0.888551647234883\\
0.2	0.879650993659419\\
0.1	0.875098867685222\\
0	0.872967837362261\\
0	1.06628216263774\\
0.1	1.06395113231478\\
0.2	1.06099900634058\\
0.3	1.05539835276512\\
0.4	1.04741322724878\\
0.5	1.04042616166244\\
0.6	1.04402988695239\\
0.7	1.04488825230833\\
0.8	1.04386396357309\\
0.9	1.04264911978456\\
1	1.03947432074901\\
}--cycle;
\addplot [color=white!55!red, forget plot]
  table[row sep=crcr]{%
1	0.852625679250985\\
0.9	0.863700880215436\\
0.8	0.872136036426909\\
0.7	0.886561747691671\\
0.6	0.908070113047607\\
0.5	0.915823838337564\\
0.4	0.900536772751223\\
0.3	0.888551647234883\\
0.2	0.879650993659419\\
0.1	0.875098867685222\\
0	0.872967837362261\\
};
\addplot [color=white!55!red, forget plot]
  table[row sep=crcr]{%
1	1.03947432074901\\
0.9	1.04264911978456\\
0.8	1.04386396357309\\
0.7	1.04488825230833\\
0.6	1.04402988695239\\
0.5	1.04042616166244\\
0.4	1.04741322724878\\
0.3	1.05539835276512\\
0.2	1.06099900634058\\
0.1	1.06395113231478\\
0	1.06628216263774\\
};
\addplot [color=red, line width=2.0pt]
  table[row sep=crcr]{%
1	0.94605\\
0.9	0.953175\\
0.8	0.958\\
0.7	0.965725\\
0.6	0.97605\\
0.5	0.978125\\
0.4	0.973974999999999\\
0.3	0.971975\\
0.2	0.970325\\
0.1	0.969525\\
0	0.969625\\
};

\addplot[area legend, draw=none, fill=orange, fill opacity=0.15, forget plot]
table[row sep=crcr] {%
x	y\\
1	0.618301029638615\\
0.9	0.6328511666224\\
0.8	0.660154422333404\\
0.7	0.709342269265207\\
0.6	0.739400904890548\\
0.5	0.749997985845497\\
0.4	0.753163085671296\\
0.3	0.746338607542076\\
0.2	0.739363170242535\\
0.1	0.730763154721956\\
0	0.720566173845261\\
0	0.876833826154741\\
0.1	0.889036845278044\\
0.2	0.899536829757465\\
0.3	0.910061392457925\\
0.4	0.915336914328705\\
0.5	0.908102014154503\\
0.6	0.886949095109451\\
0.7	0.847857730734793\\
0.8	0.805045577666597\\
0.9	0.7680988333776\\
1	0.746648970361385\\
}--cycle;
\addplot [color=white!55!orange, forget plot]
  table[row sep=crcr]{%
1	0.618301029638615\\
0.9	0.6328511666224\\
0.8	0.660154422333404\\
0.7	0.709342269265207\\
0.6	0.739400904890548\\
0.5	0.749997985845497\\
0.4	0.753163085671296\\
0.3	0.746338607542076\\
0.2	0.739363170242535\\
0.1	0.730763154721956\\
0	0.720566173845261\\
};
\addplot [color=white!55!orange, forget plot]
  table[row sep=crcr]{%
1	0.746648970361385\\
0.9	0.7680988333776\\
0.8	0.805045577666597\\
0.7	0.847857730734793\\
0.6	0.886949095109451\\
0.5	0.908102014154503\\
0.4	0.915336914328705\\
0.3	0.910061392457925\\
0.2	0.899536829757465\\
0.1	0.889036845278044\\
0	0.876833826154741\\
};
\addplot [color=orange, line width=2.0pt]
  table[row sep=crcr]{%
1	0.682475\\
0.9	0.700475\\
0.8	0.7326\\
0.7	0.7786\\
0.6	0.813174999999999\\
0.5	0.82905\\
0.4	0.83425\\
0.3	0.8282\\
0.2	0.81945\\
0.1	0.8099\\
0	0.798700000000001\\
};

\addplot[area legend, draw=none, fill=violet, fill opacity=0.15, forget plot]
table[row sep=crcr] {%
x	y\\
1	0.449099414522259\\
0.9	0.458307405118972\\
0.8	0.47713814094441\\
0.7	0.487883166773127\\
0.6	0.495028965779751\\
0.5	0.49096870410521\\
0.4	0.485538054351463\\
0.3	0.479368185306367\\
0.2	0.472319750885385\\
0.1	0.463461774493434\\
0	0.453133826061718\\
0	0.598366173938282\\
0.1	0.609938225506567\\
0.2	0.620130249114616\\
0.3	0.629631814693632\\
0.4	0.637411945648537\\
0.5	0.641281295894791\\
0.6	0.632721034220249\\
0.7	0.605166833226873\\
0.8	0.584261859055592\\
0.9	0.568342594881029\\
1	0.554900585477741\\
}--cycle;
\addplot [color=white!55!violet, forget plot]
  table[row sep=crcr]{%
1	0.449099414522259\\
0.9	0.458307405118972\\
0.8	0.47713814094441\\
0.7	0.487883166773127\\
0.6	0.495028965779751\\
0.5	0.49096870410521\\
0.4	0.485538054351463\\
0.3	0.479368185306367\\
0.2	0.472319750885385\\
0.1	0.463461774493434\\
0	0.453133826061718\\
};
\addplot [color=white!55!violet, forget plot]
  table[row sep=crcr]{%
1	0.554900585477741\\
0.9	0.568342594881029\\
0.8	0.584261859055592\\
0.7	0.605166833226873\\
0.6	0.632721034220249\\
0.5	0.641281295894791\\
0.4	0.637411945648537\\
0.3	0.629631814693632\\
0.2	0.620130249114616\\
0.1	0.609938225506567\\
0	0.598366173938282\\
};
\addplot [color=violet, line width=2.0pt]
  table[row sep=crcr]{%
1	0.502\\
0.9	0.513325\\
0.8	0.530700000000001\\
0.7	0.546525\\
0.6	0.563875\\
0.5	0.566125\\
0.4	0.561475\\
0.3	0.5545\\
0.2	0.546225\\
0.1	0.5367\\
0	0.52575\\
};

\addplot[area legend, draw=none, fill=blue, fill opacity=0.15, forget plot]
table[row sep=crcr] {%
x	y\\
1	0.921976440414829\\
0.9	0.941628663352161\\
0.8	0.968673661921001\\
0.7	0.960083657673863\\
0.6	0.950245456592304\\
0.5	0.866892985634337\\
0.4	0.204595999210155\\
0.3	0.259982621628931\\
0.2	0.26010504113449\\
0.1	0.260053047623141\\
0	0.260118329852952\\
0	0.307231670147048\\
0.1	0.307346952376859\\
0.2	0.30724495886551\\
0.3	0.307217378371069\\
0.4	0.392454000789846\\
0.5	1.03280701436566\\
0.6	1.0306545434077\\
0.7	1.02646634232614\\
0.8	1.022576338079\\
0.9	1.03367133664784\\
1	1.03802355958517\\
}--cycle;
\addplot [color=white!55!blue,dashed, forget plot]
  table[row sep=crcr]{%
1	0.921976440414829\\
0.9	0.941628663352161\\
0.8	0.968673661921001\\
0.7	0.960083657673863\\
0.6	0.950245456592304\\
0.5	0.866892985634337\\
0.4	0.204595999210155\\
0.3	0.259982621628931\\
0.2	0.26010504113449\\
0.1	0.260053047623141\\
0	0.260118329852952\\
};
\addplot [color=white!55!blue,dashed, forget plot]
  table[row sep=crcr]{%
1	1.03802355958517\\
0.9	1.03367133664784\\
0.8	1.022576338079\\
0.7	1.02646634232614\\
0.6	1.0306545434077\\
0.5	1.03280701436566\\
0.4	0.392454000789846\\
0.3	0.307217378371069\\
0.2	0.30724495886551\\
0.1	0.307346952376859\\
0	0.307231670147048\\
};
\addplot [color=blue, dashed, line width=2.0pt, forget plot]
  table[row sep=crcr]{%
1	0.979999999999999\\
0.9	0.987649999999999\\
0.8	0.995625\\
0.7	0.993275\\
0.6	0.99045\\
0.5	0.94985\\
0.4	0.298525\\
0.3	0.2836\\
0.2	0.283675\\
0.1	0.2837\\
0	0.283675\\
};

\addplot[area legend, draw=none, fill=orange, fill opacity=0.15, forget plot]
table[row sep=crcr] {%
x	y\\
1	0.618301029638615\\
0.9	0.648384950850581\\
0.8	0.688794660136543\\
0.7	0.718459309978591\\
0.6	0.738342800257474\\
0.5	0.695744670629611\\
0.4	0.467895094762895\\
0.3	0.257344603517596\\
0.2	0.257669400959506\\
0.1	0.25830591729226\\
0	0.259752110899754\\
0	0.299347889100247\\
0.1	0.29859408270774\\
0.2	0.301380599040494\\
0.3	0.313105396482404\\
0.4	0.872204905237105\\
0.5	0.913355329370389\\
0.6	0.887807199742525\\
0.7	0.85424069002141\\
0.8	0.815855339863456\\
0.9	0.78096504914942\\
1	0.746648970361385\\
}--cycle;
\addplot [color=white!55!orange, dashed, forget plot]
  table[row sep=crcr]{%
1	0.618301029638615\\
0.9	0.648384950850581\\
0.8	0.688794660136543\\
0.7	0.718459309978591\\
0.6	0.738342800257474\\
0.5	0.695744670629611\\
0.4	0.467895094762895\\
0.3	0.257344603517596\\
0.2	0.257669400959506\\
0.1	0.25830591729226\\
0	0.259752110899754\\
};
\addplot [color=white!55!orange,dashed, forget plot]
  table[row sep=crcr]{%
1	0.746648970361385\\
0.9	0.78096504914942\\
0.8	0.815855339863456\\
0.7	0.85424069002141\\
0.6	0.887807199742525\\
0.5	0.913355329370389\\
0.4	0.872204905237105\\
0.3	0.313105396482404\\
0.2	0.301380599040494\\
0.1	0.29859408270774\\
0	0.299347889100247\\
};
\addplot [color=orange, dashed, line width=2.0pt, forget plot]
  table[row sep=crcr]{%
1	0.682475\\
0.9	0.714675\\
0.8	0.752324999999999\\
0.7	0.78635\\
0.6	0.813075\\
0.5	0.80455\\
0.4	0.67005\\
0.3	0.285225\\
0.2	0.279525\\
0.1	0.27845\\
0	0.27955\\
};

\end{axis}
\end{tikzpicture}%
\label{fig:arti3:X}}\hfill
    \subfigure[$ \mathrm{MAP}(\Psi,\Psi^\bullet)$]{
%
%
\begin{tikzpicture}

\begin{axis}[%
width=0.4\textwidth,
height=1.75in,
at={(0.758in,0.481in)},
scale only axis,
xmin=0,
xmax=1,
xlabel style={font=\color{white!15!black}},
xlabel={$\beta$},
ymin=0.4,
ymax=1.0,
axis background/.style={fill=white},
legend style={at={(0.97,0.03)}, anchor=south east, legend cell align=left, align=left, draw=white!15!black}
]

\addplot[area legend, draw=none, fill=blue, fill opacity=0.15, forget plot]
table[row sep=crcr] {%
x	y\\
1	0.846332662824924\\
0.9	0.846332662824924\\
0.8	0.846332662824924\\
0.7	0.847331082114463\\
0.6	0.852089221437251\\
0.5	0.857364885510157\\
0.4	0.858385936961506\\
0.3	0.856879801264\\
0.2	0.856328273178749\\
0.1	0.854485977885301\\
0	0.854875488574715\\
0	1.08152451142529\\
0.1	1.0817540221147\\
0.2	1.08071172682125\\
0.3	1.080320198736\\
0.4	1.07961406303849\\
0.5	1.08007511448984\\
0.6	1.08239077856275\\
0.7	1.08506891788554\\
0.8	1.08558733717508\\
0.9	1.08558733717508\\
1	1.08558733717508\\
}--cycle;
\addplot [color=white!55!blue, forget plot]
  table[row sep=crcr]{%
1	0.846332662824924\\
0.9	0.846332662824924\\
0.8	0.846332662824924\\
0.7	0.847331082114463\\
0.6	0.852089221437251\\
0.5	0.857364885510157\\
0.4	0.858385936961506\\
0.3	0.856879801264\\
0.2	0.856328273178749\\
0.1	0.854485977885301\\
0	0.854875488574715\\
};
\addplot [color=white!55!blue, forget plot]
  table[row sep=crcr]{%
1	1.08558733717508\\
0.9	1.08558733717508\\
0.8	1.08558733717508\\
0.7	1.08506891788554\\
0.6	1.08239077856275\\
0.5	1.08007511448984\\
0.4	1.07961406303849\\
0.3	1.080320198736\\
0.2	1.08071172682125\\
0.1	1.0817540221147\\
0	1.08152451142529\\
};
\addplot [color=blue, line width=2.0pt]
  table[row sep=crcr]{%
1	0.96596\\
0.9	0.96596\\
0.8	0.96596\\
0.7	0.9662\\
0.6	0.96724\\
0.5	0.96872\\
0.4	0.969\\
0.3	0.9686\\
0.2	0.96852\\
0.1	0.96812\\
0	0.9682\\
};
\addlegendentry{$\varepsilon=0.0$}

\addplot[area legend, draw=none, fill=red, fill opacity=0.15, forget plot]
table[row sep=crcr] {%
x	y\\
1	0.884462334679023\\
0.9	0.884004423834752\\
0.8	0.884462334679023\\
0.7	0.884691879236547\\
0.6	0.886096166504468\\
0.5	0.887940546084033\\
0.4	0.887826673873887\\
0.3	0.883623301677836\\
0.2	0.884487604566604\\
0.1	0.885059002339384\\
0	0.88573483430123\\
0	1.07402516569877\\
0.1	1.07446099766062\\
0.2	1.0747923954334\\
0.3	1.07533669832216\\
0.4	1.07265332612611\\
0.5	1.07261945391597\\
0.6	1.07382383349553\\
0.7	1.07466812076345\\
0.8	1.07481766532098\\
0.9	1.07511557616525\\
1	1.07481766532098\\
}--cycle;
\addplot [color=white!55!red, forget plot]
  table[row sep=crcr]{%
1	0.884462334679023\\
0.9	0.884004423834752\\
0.8	0.884462334679023\\
0.7	0.884691879236547\\
0.6	0.886096166504468\\
0.5	0.887940546084033\\
0.4	0.887826673873887\\
0.3	0.883623301677836\\
0.2	0.884487604566604\\
0.1	0.885059002339384\\
0	0.88573483430123\\
};
\addplot [color=white!55!red, forget plot]
  table[row sep=crcr]{%
1	1.07481766532098\\
0.9	1.07511557616525\\
0.8	1.07481766532098\\
0.7	1.07466812076345\\
0.6	1.07382383349553\\
0.5	1.07261945391597\\
0.4	1.07265332612611\\
0.3	1.07533669832216\\
0.2	1.0747923954334\\
0.1	1.07446099766062\\
0	1.07402516569877\\
};
\addplot [color=red, line width=2.0pt]
  table[row sep=crcr]{%
1	0.97964\\
0.9	0.97956\\
0.8	0.97964\\
0.7	0.97968\\
0.6	0.97996\\
0.5	0.98028\\
0.4	0.98024\\
0.3	0.97948\\
0.2	0.97964\\
0.1	0.97976\\
0	0.97988\\
};
\addlegendentry{$\varepsilon=0.5$}

\addplot[area legend, draw=none, fill=orange, fill opacity=0.15, forget plot]
table[row sep=crcr] {%
x	y\\
1	0.91827735413679\\
0.9	0.918856948844833\\
0.8	0.918802505426541\\
0.7	0.925466143469988\\
0.6	0.924591243644197\\
0.5	0.924307732516831\\
0.4	0.924484673287666\\
0.3	0.925060065611219\\
0.2	0.924680273914746\\
0.1	0.922850612778264\\
0	0.92338386864799\\
0	1.05397613135201\\
0.1	1.05498938722174\\
0.2	1.05395972608526\\
0.3	1.05381993438878\\
0.4	1.05423532671233\\
0.5	1.05465226748317\\
0.6	1.0543687563558\\
0.7	1.05365385653001\\
0.8	1.05703749457346\\
0.9	1.05714305115517\\
1	1.05724264586321\\
}--cycle;
\addplot [color=white!55!orange, forget plot]
  table[row sep=crcr]{%
1	0.91827735413679\\
0.9	0.918856948844833\\
0.8	0.918802505426541\\
0.7	0.925466143469988\\
0.6	0.924591243644197\\
0.5	0.924307732516831\\
0.4	0.924484673287666\\
0.3	0.925060065611219\\
0.2	0.924680273914746\\
0.1	0.922850612778264\\
0	0.92338386864799\\
};
\addplot [color=white!55!orange, forget plot]
  table[row sep=crcr]{%
1	1.05724264586321\\
0.9	1.05714305115517\\
0.8	1.05703749457346\\
0.7	1.05365385653001\\
0.6	1.0543687563558\\
0.5	1.05465226748317\\
0.4	1.05423532671233\\
0.3	1.05381993438878\\
0.2	1.05395972608526\\
0.1	1.05498938722174\\
0	1.05397613135201\\
};
\addplot [color=orange, line width=2.0pt]
  table[row sep=crcr]{%
1	0.987760000000001\\
0.9	0.988000000000001\\
0.8	0.98792\\
0.7	0.989560000000001\\
0.6	0.98948\\
0.5	0.98948\\
0.4	0.98936\\
0.3	0.989440000000001\\
0.2	0.989320000000001\\
0.1	0.988920000000001\\
0	0.988680000000001\\
};
\addlegendentry{$\varepsilon=0.7$}

\addplot[area legend, draw=none, fill=violet, fill opacity=0.15, forget plot]
table[row sep=crcr] {%
x	y\\
1	0.694425804204209\\
0.9	0.724382600055435\\
0.8	0.745030898536485\\
0.7	0.752551482756622\\
0.6	0.759906855483498\\
0.5	0.760233314599881\\
0.4	0.760567376282041\\
0.3	0.759373341901953\\
0.2	0.757269317352369\\
0.1	0.753310134573722\\
0	0.742107972239472\\
0	0.952772027760525\\
0.1	0.958769865426276\\
0.2	0.961690682647629\\
0.3	0.963106658098044\\
0.4	0.963432623717955\\
0.5	0.963366685400115\\
0.6	0.965533144516499\\
0.7	0.965128517243375\\
0.8	0.965049101463512\\
0.9	0.961137399944563\\
1	0.950374195795789\\
}--cycle;
\addplot [color=white!55!violet, forget plot]
  table[row sep=crcr]{%
1	0.694425804204209\\
0.9	0.724382600055435\\
0.8	0.745030898536485\\
0.7	0.752551482756622\\
0.6	0.759906855483498\\
0.5	0.760233314599881\\
0.4	0.760567376282041\\
0.3	0.759373341901953\\
0.2	0.757269317352369\\
0.1	0.753310134573722\\
0	0.742107972239472\\
};
\addplot [color=white!55!violet, forget plot]
  table[row sep=crcr]{%
1	0.950374195795789\\
0.9	0.961137399944563\\
0.8	0.965049101463512\\
0.7	0.965128517243375\\
0.6	0.965533144516499\\
0.5	0.963366685400115\\
0.4	0.963432623717955\\
0.3	0.963106658098044\\
0.2	0.961690682647629\\
0.1	0.958769865426276\\
0	0.952772027760525\\
};
\addplot [color=violet, line width=2.0pt]
  table[row sep=crcr]{%
1	0.822399999999999\\
0.9	0.842759999999999\\
0.8	0.855039999999998\\
0.7	0.858839999999998\\
0.6	0.862719999999998\\
0.5	0.861799999999998\\
0.4	0.861999999999998\\
0.3	0.861239999999999\\
0.2	0.859479999999999\\
0.1	0.856039999999999\\
0	0.847439999999999\\
};
\addlegendentry{$\varepsilon=0.8$}

\end{axis}
\end{tikzpicture}%
\label{fig:arti3:Y}} 
	\end{minipage}
    \rulesep 
	\begin{minipage}{0.3\textwidth}
	\begin{center}
	 \subfigure[$k=3$]{\includegraphics[width=0.6\textwidth]{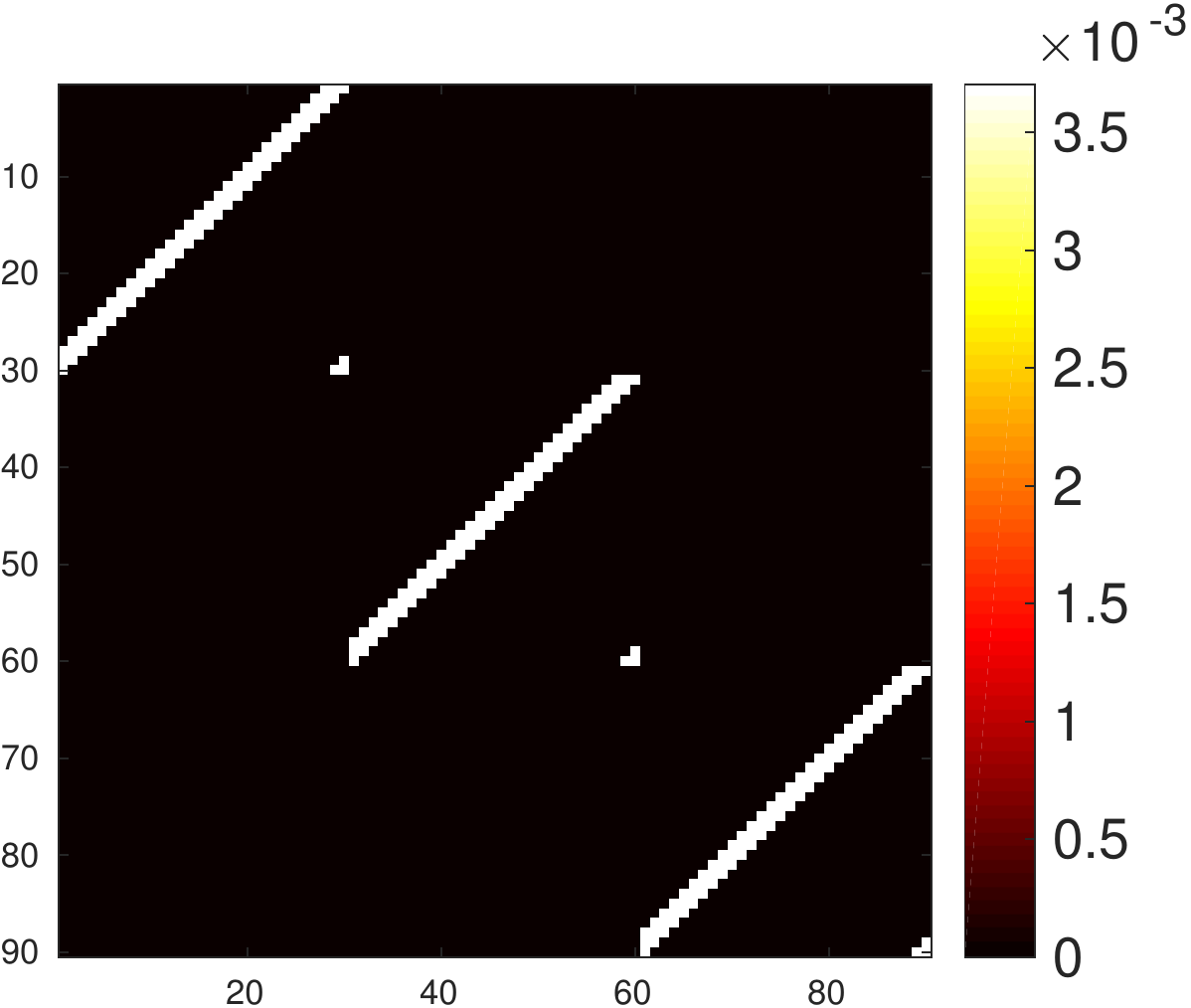}\label{fig:arti3:k3}}
     \subfigure[$k=15$]{\includegraphics[width=0.6\textwidth]{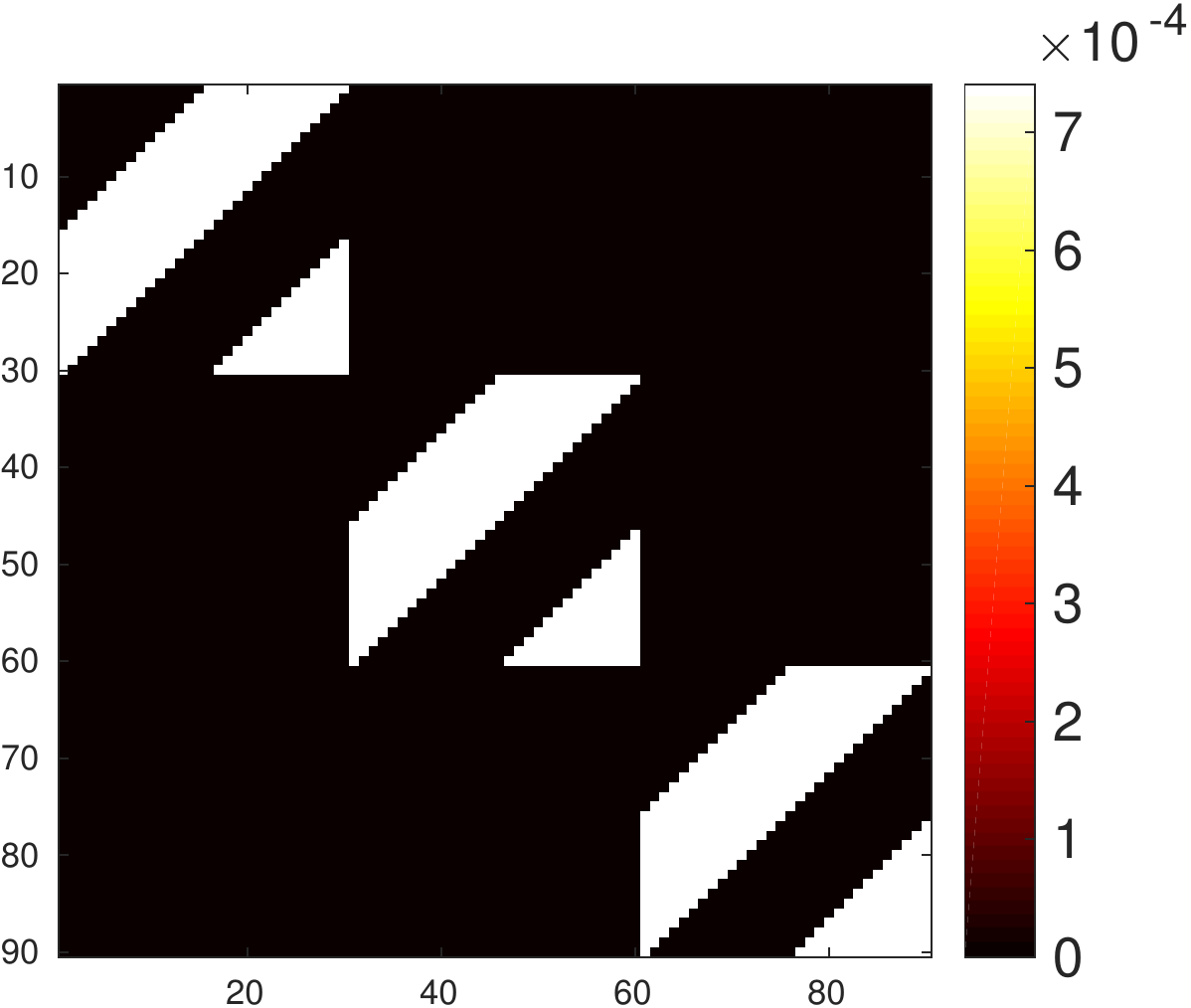}\label{fig:arti3:k15}}
\end{center}
	 \subfigure[$ \mathrm{MAP}(\Phi,\Phi^\bullet)$]{
%
%

\begin{tikzpicture}

\begin{axis}[%
width=0.8\textwidth,
height=1.75in,
at={(0.758in,0.481in)},
scale only axis,
xmin=0,
xmax=1,
xlabel style={font=\color{white!15!black}},
xlabel={$\beta$},
ymin=0,
ymax=1,
axis background/.style={fill=white},
legend style={at={(0.97,0.03)}, anchor=south east, legend cell align=left, align=left, draw=white!15!black}
]

\addplot[area legend, draw=none, fill=blue, fill opacity=0.15, forget plot]
table[row sep=crcr] {%
x	y\\
1	0.433285133177592\\
0.9	0.475322512431909\\
0.8	0.476153668167523\\
0.7	0.476153668167523\\
0.6	0.476153668167523\\
0.5	0.476153668167523\\
0.4	0.476153668167523\\
0.3	0.476153668167523\\
0.2	0.476153668167523\\
0.1	0.476153668167523\\
0	0.476153668167523\\
0	0.700735220721367\\
0.1	0.700735220721367\\
0.2	0.700735220721367\\
0.3	0.700735220721367\\
0.4	0.700735220721367\\
0.5	0.700735220721367\\
0.6	0.700735220721367\\
0.7	0.700735220721367\\
0.8	0.700735220721367\\
0.9	0.697210820901426\\
1	0.57484820015574\\
}--cycle;
\addplot [color=white!55!blue, forget plot]
  table[row sep=crcr]{%
1	0.433285133177592\\
0.9	0.475322512431909\\
0.8	0.476153668167523\\
0.7	0.476153668167523\\
0.6	0.476153668167523\\
0.5	0.476153668167523\\
0.4	0.476153668167523\\
0.3	0.476153668167523\\
0.2	0.476153668167523\\
0.1	0.476153668167523\\
0	0.476153668167523\\
};
\addplot [color=white!55!blue, forget plot]
  table[row sep=crcr]{%
1	0.57484820015574\\
0.9	0.697210820901426\\
0.8	0.700735220721367\\
0.7	0.700735220721367\\
0.6	0.700735220721367\\
0.5	0.700735220721367\\
0.4	0.700735220721367\\
0.3	0.700735220721367\\
0.2	0.700735220721367\\
0.1	0.700735220721367\\
0	0.700735220721367\\
};
\addplot [color=blue, line width=2.0pt]
  table[row sep=crcr]{%
1	0.504066666666666\\
0.9	0.586266666666667\\
0.8	0.588444444444445\\
0.7	0.588444444444445\\
0.6	0.588444444444445\\
0.5	0.588444444444445\\
0.4	0.588444444444445\\
0.3	0.588444444444445\\
0.2	0.588444444444445\\
0.1	0.588444444444445\\
0	0.588444444444445\\
};
\addlegendentry{$k=3$}

\addplot[area legend, draw=none, fill=blue, fill opacity=0.15, forget plot]
table[row sep=crcr] {%
x	y\\
1	0.433285133177592\\
0.9	0.451960159979409\\
0.8	0.453804495473001\\
0.7	0.442636794457567\\
0.6	0.429047363767927\\
0.5	0.40689406104417\\
0.4	0.390798618086938\\
0.3	0.352250490208822\\
0.2	0.342118915198903\\
0.1	0.341894285376762\\
0	0.342023097435285\\
0	0.360154680342491\\
0.1	0.360150159067681\\
0.2	0.368547751467762\\
0.3	0.424771732013397\\
0.4	0.489512493024172\\
0.5	0.520883716733606\\
0.6	0.569930414009852\\
0.7	0.601985427764655\\
0.8	0.618995504526999\\
0.9	0.663017617798369\\
1	0.57484820015574\\
}--cycle;
\addplot [color=white!55!blue,dashed,  forget plot]
  table[row sep=crcr]{%
1	0.433285133177592\\
0.9	0.451960159979409\\
0.8	0.453804495473001\\
0.7	0.442636794457567\\
0.6	0.429047363767927\\
0.5	0.40689406104417\\
0.4	0.390798618086938\\
0.3	0.352250490208822\\
0.2	0.342118915198903\\
0.1	0.341894285376762\\
0	0.342023097435285\\
};
\addplot [color=white!55!blue,dashed,  forget plot]
  table[row sep=crcr]{%
1	0.57484820015574\\
0.9	0.663017617798369\\
0.8	0.618995504526999\\
0.7	0.601985427764655\\
0.6	0.569930414009852\\
0.5	0.520883716733606\\
0.4	0.489512493024172\\
0.3	0.424771732013397\\
0.2	0.368547751467762\\
0.1	0.360150159067681\\
0	0.360154680342491\\
};
\addplot [color=blue, dashed, line width=2.0pt, forget plot]
  table[row sep=crcr]{%
1	0.504066666666666\\
0.9	0.557488888888889\\
0.8	0.5364\\
0.7	0.522311111111111\\
0.6	0.499488888888889\\
0.5	0.463888888888888\\
0.4	0.440155555555555\\
0.3	0.38851111111111\\
0.2	0.355333333333332\\
0.1	0.351022222222221\\
0	0.351088888888888\\
};

\addplot[area legend, draw=none, fill=violet, fill opacity=0.15, forget plot]
table[row sep=crcr] {%
x	y\\
1	0.668656875169\\
0.9	0.833033918863199\\
0.8	0.850866894855301\\
0.7	0.876646883123921\\
0.6	0.876646883123921\\
0.5	0.876646883123921\\
0.4	0.876646883123921\\
0.3	0.876646883123921\\
0.2	0.876646883123921\\
0.1	0.876646883123921\\
0	0.876646883123921\\
0	1.06206422798719\\
0.1	1.06206422798719\\
0.2	1.06206422798719\\
0.3	1.06206422798719\\
0.4	1.06206422798719\\
0.5	1.06206422798719\\
0.6	1.06206422798719\\
0.7	1.06206422798719\\
0.8	1.06668866070026\\
0.9	1.0677660811368\\
1	0.993432013719888\\
}--cycle;
\addplot [color=white!55!violet, forget plot]
  table[row sep=crcr]{%
1	0.668656875169\\
0.9	0.833033918863199\\
0.8	0.850866894855301\\
0.7	0.876646883123921\\
0.6	0.876646883123921\\
0.5	0.876646883123921\\
0.4	0.876646883123921\\
0.3	0.876646883123921\\
0.2	0.876646883123921\\
0.1	0.876646883123921\\
0	0.876646883123921\\
};
\addplot [color=white!55!violet, forget plot]
  table[row sep=crcr]{%
1	0.993432013719888\\
0.9	1.0677660811368\\
0.8	1.06668866070026\\
0.7	1.06206422798719\\
0.6	1.06206422798719\\
0.5	1.06206422798719\\
0.4	1.06206422798719\\
0.3	1.06206422798719\\
0.2	1.06206422798719\\
0.1	1.06206422798719\\
0	1.06206422798719\\
};
\addplot [color=violet, line width=2.0pt]
  table[row sep=crcr]{%
1	0.831044444444444\\
0.9	0.9504\\
0.8	0.958777777777778\\
0.7	0.969355555555556\\
0.6	0.969355555555556\\
0.5	0.969355555555556\\
0.4	0.969355555555556\\
0.3	0.969355555555556\\
0.2	0.969355555555556\\
0.1	0.969355555555556\\
0	0.969355555555556\\
};
\addlegendentry{$k=6$}

\addplot[area legend, draw=none, fill=violet, fill opacity=0.15, forget plot]
table[row sep=crcr] {%
x	y\\
1	0.668656875169\\
0.9	0.841132978826642\\
0.8	0.775070281496503\\
0.7	0.667204438158594\\
0.6	0.573710446443172\\
0.5	0.44592502428489\\
0.4	0.400152288698829\\
0.3	0.338960344628178\\
0.2	0.337811923376835\\
0.1	0.337464700855064\\
0	0.33789026034865\\
0	0.351087517429126\\
0.1	0.350757521367156\\
0.2	0.351121409956496\\
0.3	0.352284099816264\\
0.4	0.565625489078947\\
0.5	0.642341642381776\\
0.6	0.883667331334604\\
0.7	1.00488445073029\\
0.8	1.05448527405905\\
0.9	1.06313368784002\\
1	0.993432013719888\\
}--cycle;
\addplot [color=white!55!violet,dashed,  forget plot]
  table[row sep=crcr]{%
1	0.668656875169\\
0.9	0.841132978826642\\
0.8	0.775070281496503\\
0.7	0.667204438158594\\
0.6	0.573710446443172\\
0.5	0.44592502428489\\
0.4	0.400152288698829\\
0.3	0.338960344628178\\
0.2	0.337811923376835\\
0.1	0.337464700855064\\
0	0.33789026034865\\
};
\addplot [color=white!55!violet,dashed,  forget plot]
  table[row sep=crcr]{%
1	0.993432013719888\\
0.9	1.06313368784002\\
0.8	1.05448527405905\\
0.7	1.00488445073029\\
0.6	0.883667331334604\\
0.5	0.642341642381776\\
0.4	0.565625489078947\\
0.3	0.352284099816264\\
0.2	0.351121409956496\\
0.1	0.350757521367156\\
0	0.351087517429126\\
};
\addplot [color=violet, dashed, line width=2.0pt, forget plot]
  table[row sep=crcr]{%
1	0.831044444444444\\
0.9	0.952133333333333\\
0.8	0.914777777777778\\
0.7	0.836044444444444\\
0.6	0.728688888888888\\
0.5	0.544133333333333\\
0.4	0.482888888888888\\
0.3	0.345622222222221\\
0.2	0.344466666666666\\
0.1	0.34411111111111\\
0	0.344488888888888\\
};

\addplot[area legend, draw=none, fill=orange, fill opacity=0.15, forget plot]
table[row sep=crcr] {%
x	y\\
1	1\\
0.9	1\\
0.8	1\\
0.7	1\\
0.6	1\\
0.5	1\\
0.4	1\\
0.3	1\\
0.2	1\\
0.1	1\\
0	1\\
0	1\\
0.1	1\\
0.2	1\\
0.3	1\\
0.4	1\\
0.5	1\\
0.6	1\\
0.7	1\\
0.8	1\\
0.9	1\\
1	1\\
}--cycle;
\addplot [color=white!55!orange, forget plot]
  table[row sep=crcr]{%
1	1\\
0.9	1\\
0.8	1\\
0.7	1\\
0.6	1\\
0.5	1\\
0.4	1\\
0.3	1\\
0.2	1\\
0.1	1\\
0	1\\
};
\addplot [color=white!55!orange, forget plot]
  table[row sep=crcr]{%
1	1\\
0.9	1\\
0.8	1\\
0.7	1\\
0.6	1\\
0.5	1\\
0.4	1\\
0.3	1\\
0.2	1\\
0.1	1\\
0	1\\
};
\addplot [color=orange, line width=2.0pt]
  table[row sep=crcr]{%
1	1\\
0.9	1\\
0.8	1\\
0.7	1\\
0.6	1\\
0.5	1\\
0.4	1\\
0.3	1\\
0.2	1\\
0.1	1\\
0	1\\
};
\addlegendentry{$k=15$}

\addplot[area legend, draw=none, fill=orange, fill opacity=0.15, forget plot]
table[row sep=crcr] {%
x	y\\
1	1\\
0.9	1\\
0.8	1\\
0.7	1\\
0.6	0.984426213483335\\
0.5	0.712276336436658\\
0.4	0.24531520315292\\
0.3	0.332054619225385\\
0.2	0.332337795644982\\
0.1	0.332115547102724\\
0	0.332337795644982\\
0	0.334551093243908\\
0.1	0.334906675119499\\
0.2	0.334551093243908\\
0.3	0.335012047441283\\
0.4	0.494240352402637\\
0.5	1.03803477467445\\
0.6	1.01424045318333\\
0.7	1\\
0.8	1\\
0.9	1\\
1	1\\
}--cycle;
\addplot [color=white!55!orange,dashed,  forget plot]
  table[row sep=crcr]{%
1	1\\
0.9	1\\
0.8	1\\
0.7	1\\
0.6	0.984426213483335\\
0.5	0.712276336436658\\
0.4	0.24531520315292\\
0.3	0.332054619225385\\
0.2	0.332337795644982\\
0.1	0.332115547102724\\
0	0.332337795644982\\
};
\addplot [color=white!55!orange, dashed, forget plot]
  table[row sep=crcr]{%
1	1\\
0.9	1\\
0.8	1\\
0.7	1\\
0.6	1.01424045318333\\
0.5	1.03803477467445\\
0.4	0.494240352402637\\
0.3	0.335012047441283\\
0.2	0.334551093243908\\
0.1	0.334906675119499\\
0	0.334551093243908\\
};
\addplot [color=orange, dashed, line width=2.0pt, forget plot]
  table[row sep=crcr]{%
1	1\\
0.9	1\\
0.8	1\\
0.7	1\\
0.6	0.999333333333333\\
0.5	0.875155555555556\\
0.4	0.369777777777779\\
0.3	0.333533333333334\\
0.2	0.333444444444445\\
0.1	0.333511111111112\\
0	0.333444444444445\\
};
\end{axis}
\end{tikzpicture}%
\label{fig:arti3:X_blocks}}
	\end{minipage}
    \caption{\rev{(a)-(c), (f)-(g): Colorplots of $P_{X,Y}$ for different noise levels $\varepsilon$ and different parameters $k$. It can be seen that the true cluster structure becomes less obvious with increasing noise levels. (d), (e), (h): Micro-averaged precision curves show the average over 500 random experiments (center line) and the standard deviation (shaded area). Solid curves correspond to \textsc{AnnITCC}, dashed curves to \textsc{sGITCC}. See text for details.}}
    \label{arti3} 
\end{figure*}

We repeated the whole procedure for 500 different probability matrices $N$. The MAP values, averaged over these 500 runs, are reported in Fig.~\ref{fig:arti3:X} and~\ref{fig:arti3:Y} (solid lines). First of all, it can be seen that even in the noiseless case, the clusters are not always identified correctly. \rev{Since we identified the correct clusters in over 90\% of the simulation runs, we believe that this effect can be explained by the algorithm getting stuck in a local optimum}. Second, one can observe the natural effect that large noise levels lead to lower MAP values -- interestingly, though, co-clustering seems to be quite robust to noise, as the MAP values in this experiment seem to decrease significantly only for $\varepsilon>0.5$, \rev{i.e., when noise starts to dominate the data matrix}. Finally, for large noise levels, it turns out that the \rev{intermediate values of $\beta$ perform better. The performance drop for larger values of $\beta$ is not due to the optimization heuristic getting stuck in bad local optima: We found that the cost of the co-clustering solution found by \textsc{AnnITCC} for large $\beta$ is lower than the cost of the ground truth.} Rather, the reason is that for $\beta=1$ the clustering solutions are uncoupled, i.e., the relevant RV for clustering rows is the noisy column RV. For a certain amount of coupling, i.e., for intermediate values of $\beta$, the relevant RV for clustering rows is more strongly related to the column \emph{clusters}, in which noise is reduced due to the averaging effect of clustering. Performance drops again when decreasing $\beta$ further; the reason is the inherent shortcoming of $\mathcal{L}_0(\Phi, \Psi)$ which is discussed at the end of Section~\ref{sec:specialAlgos} and in~\cite{Amjad_GeneralizedMA}.

\rev{The second experiment investigates the effect of intra-cluster coupling between $X$ and $Y$. We choose $|\dom{X}|= |\dom{Y}|=90$ and $|\overline{\dom{X}}| = |\overline{\dom{Y}}|=3$ to avoid the effects discussed in Sec.~\ref{sec:differentCardinalities} and generate a joint probability distribution
\begin{equation}\label{eq:circulant}
P_{X,Y} = \left[\begin{array}{ccc}
	\mathbf{C} & 0 & 0\\ 0 &\mathbf{C} & 0 \\ 0 & 0 & \mathbf{C}
\end{array}\right]
\end{equation}
where $\mathbf{C}$ is a $30\times 30$ circulant matrix the first row of which consists of $30-k$ zeros followed by $k$ entries equal to $\frac{1}{k|\dom{X}|}$. Each subsequent row of $\mathbf{C}$ is obtained by a circular shift of the previous row. Fig.~\ref{fig:arti3:k3} and Fig.~\ref{fig:arti3:k15} show $P_{X,Y}$ for $k=3$ and $k=15$, respectively. The ground truth co-clustering is given by the block structure of $P_{X,Y}$.}

\rev{It is clear that, as $k$ decreases, the intra-cluster coupling between $X$ and $Y$ increases. To see this note that, for $k=30$, $X$ does not contain more information about $Y$ than the ground truth cluster $\overline{X}$ does, whereas for $k=1$, $X$ specifies $Y$ uniquely. Fig.~\ref{fig:arti3:X_blocks} shows the average MAP values obtained by running \textsc{AnnITCC} $500$ times with random initializations. Since the experimental setup is symmetric we only show the results for $\Phi$.  First, we observe that with decreasing $k$ the performance deteriorates. This is intuitive considering that with decreasing $k$ the clustering structure becomes less obvious. For $k=30$, $P_{X,Y}$ is uniform in the the blocks whereas for $k=1$, the colums of $P_{X,Y}$ can be reordered such that $P_{X,Y}$ is a diagonal matrix with no clear co-clustering structure. Second, $\beta=1$ does not lead to the best results for increased coupling, despite the fact that the global optimum of $\mathcal{L}_1$ coincides with the ground truth.  Apparently, the optimization heuristic tends to terminate in poor local optima more often for $\beta=1$ than for smaller values of $\beta$. This is because for $\beta=1$ the two clustering solutions are decoupled, i.e., $\Phi$ and $\Psi$ are determined independently of each other, while smaller $\beta$ explicitly assumes coupled clusterings. We thus conclude that smaller values of $\beta$ detect intra-cluster coupled co-clusters more robustly.}


\rev{Finally we noticed that for both synthetic datasets, the MAP curves are relatively flat in many scenarios. One may think that this is due to \textsc{AnnITCC} getting stuck in a local optimum for a certain $\beta$, which it is not able to escape from for the subsequent lower $\beta$ values. This is not the case: Figs.~\ref{fig:arti3:X_blocks} and~\ref{fig:arti3:X} show that the results obtained by running \textsc{sGITCC} (dotted lines) are almost identical to those obtained from \textsc{AnnITCC} for larger values of $\beta$ until where both of them reach the peak performance. Subsequently, for smaller values of $\beta$, the performance of \textsc{sGITCC} dropped significantly due to the reasons outlined at the end of Section~\ref{sec:specialAlgos}, justifying using \textsc{AnnITCC} for these values of $\beta$.}

\subsection{Guiding Principles for Choosing $\beta$}
\rev{Although in this paper we do not propose a heuristic to find the suitable value (or range) of $\beta$ for a given dataset, the examples and experiments in this section admit providing the following guiding principles: }
\rev{\begin{itemize}
	\item For large differences between target cardinalities $|\overline{\dom{X}}|$ and $|\overline{\dom{Y}}|$, larger values of $\beta$ may lead to better results due to the increasingly decoupled nature of the cost function for increasing $\beta$. 
	\item For datasets with highly imbalanced (co-)clusters, smaller values of $\beta$ are more suitable (but only when one can manage to avoid optimization issues linked to smaller values of $\beta$).
	\item In general, co-clustering using $\mathcal{L}_\beta$ and $\beta$-annealing seems to be robust to noise. For large noise levels, however, intermediate values of $\beta$ tend to perform better due to noise averaging.
	\item In presence of intra-cluster coupling, local optima of $\mathcal{L}_\beta$ are more prominent for $\beta$ close to $1$. The correct co-clusterings are found more robustly for intermediate values of $\beta$. 
\end{itemize}}

\section{Real-World  Experiments}\label{sec:experiments}

\subsection{Document Classification by Co-Clustering of Words and Documents - Newsgroup20 Data Set}\label{sec:NG20}
\subsubsection{Dataset, Preprocessing, and Simulation Settings}

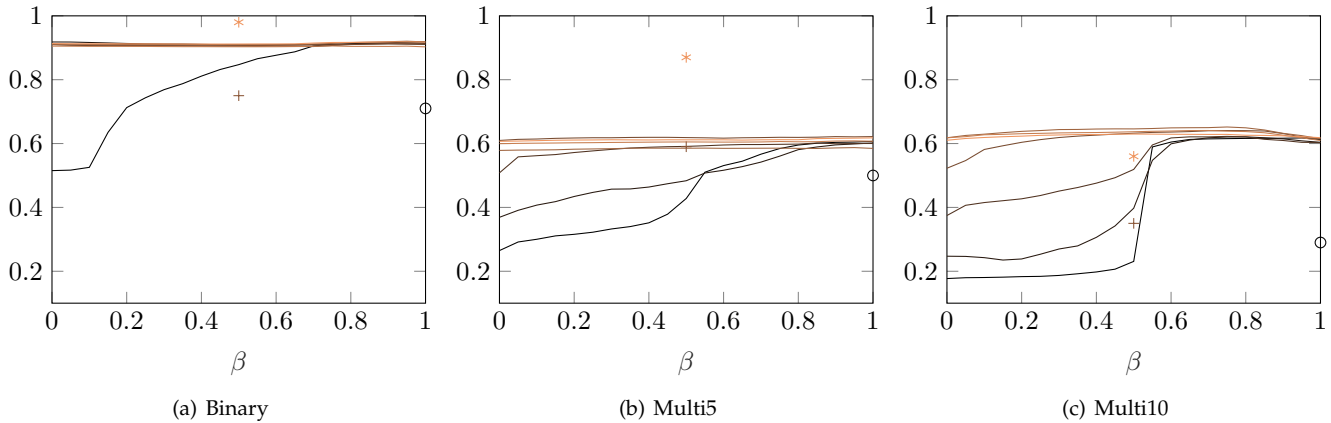
\begin{figure*}[t]
   \centering 
   \subfigure[Binary]{
%
%
\definecolor{mycolor1}{rgb}{0.15625,0.09765,0.06219}%
\definecolor{mycolor2}{rgb}{0.31250,0.19530,0.12437}%
\definecolor{mycolor3}{rgb}{0.46875,0.29295,0.18656}%
\definecolor{mycolor4}{rgb}{0.62500,0.39060,0.24875}%
\definecolor{mycolor5}{rgb}{0.78125,0.48825,0.31094}%
\definecolor{mycolor6}{rgb}{0.93750,0.58590,0.37312}%
\begin{tikzpicture}

\begin{axis}[%
width=0.27\textwidth,
height=1.5in,
at={(0.758in,0.481in)},
scale only axis,
xmin=0,
xmax=1,
xlabel style={font=\color{white!15!black}},
xlabel={$\beta$},
ymin=0.1,
ymax=1,
axis background/.style={fill=white}
]
\addplot [color=black, forget plot]
  table[row sep=crcr]{%
1	0.9185\\
0.95	0.918\\
0.9	0.9154\\
0.85	0.9141\\
0.8	0.9128\\
0.75	0.911\\
0.7	0.9057\\
0.65	0.8873\\
0.6	0.8766\\
0.55	0.8657\\
0.5	0.8472\\
0.45	0.8319\\
0.4	0.8112\\
0.35	0.7876\\
0.3	0.7688\\
0.25	0.7433\\
0.2	0.7126\\
0.15	0.635\\
0.0999999999999997	0.5255\\
0.0499999999999997	0.5168\\
-3.19189119579733e-16	0.5152\\
};
\addplot [color=mycolor1, forget plot]
  table[row sep=crcr]{%
1	0.9137\\
0.95	0.9155\\
0.9	0.9154\\
0.85	0.9142\\
0.8	0.9109\\
0.75	0.9094\\
0.7	0.9098\\
0.65	0.9097\\
0.6	0.9098\\
0.55	0.9087\\
0.5	0.909\\
0.45	0.9097\\
0.4	0.9106\\
0.35	0.9116\\
0.3	0.9118\\
0.25	0.9124\\
0.2	0.9135\\
0.15	0.9152\\
0.0999999999999997	0.9166\\
0.0499999999999997	0.918\\
-3.19189119579733e-16	0.9182\\
};
\addplot [color=mycolor2, forget plot]
  table[row sep=crcr]{%
1	0.9106\\
0.95	0.9113\\
0.9	0.9123\\
0.85	0.9114\\
0.8	0.9101\\
0.75	0.9072\\
0.7	0.9039\\
0.65	0.9041\\
0.6	0.9053\\
0.55	0.9047\\
0.5	0.905\\
0.45	0.905\\
0.4	0.9054\\
0.35	0.9061\\
0.3	0.9063\\
0.25	0.9064\\
0.2	0.9064\\
0.15	0.9066\\
0.0999999999999997	0.9069\\
0.0499999999999997	0.9078\\
-3.19189119579733e-16	0.91\\
};
\addplot [color=mycolor3, forget plot]
  table[row sep=crcr]{%
1	0.9123\\
0.95	0.913\\
0.9	0.9127\\
0.85	0.9116\\
0.8	0.911\\
0.75	0.9093\\
0.7	0.9088\\
0.65	0.9082\\
0.6	0.9078\\
0.55	0.907\\
0.5	0.9076\\
0.45	0.9078\\
0.4	0.9079\\
0.35	0.908\\
0.3	0.908\\
0.25	0.9079\\
0.2	0.9077\\
0.15	0.9084\\
0.0999999999999997	0.9088\\
0.0499999999999997	0.9092\\
-3.19189119579733e-16	0.9103\\
};
\addplot [color=mycolor4, forget plot]
  table[row sep=crcr]{%
1	0.9184\\
0.95	0.9206\\
0.9	0.919\\
0.85	0.9175\\
0.8	0.9163\\
0.75	0.9136\\
0.7	0.912\\
0.65	0.909\\
0.6	0.9093\\
0.55	0.9095\\
0.5	0.9097\\
0.45	0.9099\\
0.4	0.9099\\
0.35	0.9099\\
0.3	0.9099\\
0.25	0.91\\
0.2	0.9102\\
0.15	0.9105\\
0.0999999999999997	0.9104\\
0.0499999999999997	0.9109\\
-3.19189119579733e-16	0.9116\\
};
\addplot [color=mycolor5, forget plot]
  table[row sep=crcr]{%
1	0.9029\\
0.95	0.9051\\
0.9	0.9042\\
0.85	0.9044\\
0.8	0.9044\\
0.75	0.9057\\
0.7	0.9046\\
0.65	0.9038\\
0.6	0.9033\\
0.55	0.9032\\
0.5	0.9032\\
0.45	0.9033\\
0.4	0.9035\\
0.35	0.9035\\
0.3	0.9038\\
0.25	0.9042\\
0.2	0.9044\\
0.15	0.9045\\
0.0999999999999997	0.9045\\
0.0499999999999997	0.9049\\
-3.19189119579733e-16	0.9052\\
};
\addplot [color=mycolor6, forget plot]
  table[row sep=crcr]{%
1	0.9179\\
0.95	0.9184\\
0.9	0.9183\\
0.85	0.9179\\
0.8	0.9169\\
0.75	0.9166\\
0.7	0.9148\\
0.65	0.9124\\
0.6	0.912\\
0.55	0.912\\
0.5	0.9118\\
0.45	0.9117\\
0.4	0.9118\\
0.35	0.912\\
0.3	0.9123\\
0.25	0.9124\\
0.2	0.9125\\
0.15	0.9125\\
0.0999999999999997	0.9125\\
0.0499999999999997	0.9126\\
-3.19189119579733e-16	0.913\\
};
\addplot [color=mycolor6, draw=none, mark=asterisk, mark options={solid, mycolor6}, forget plot]
  table[row sep=crcr]{%
0.5	0.98\\
};
\addplot [color=black!30!mycolor5, draw=none, mark=+, mark options={solid, black!30!mycolor5}, forget plot]
  table[row sep=crcr]{%
0.5	0.75\\
};
\addplot [color=black, draw=none, mark=o, mark options={solid, black}, forget plot]
  table[row sep=crcr]{%
1	0.71\\
};
\end{axis}
\end{tikzpicture}%
} 
   \subfigure[Multi5]{
%
%
\definecolor{mycolor1}{rgb}{0.15625,0.09765,0.06219}%
\definecolor{mycolor2}{rgb}{0.31250,0.19530,0.12437}%
\definecolor{mycolor3}{rgb}{0.46875,0.29295,0.18656}%
\definecolor{mycolor4}{rgb}{0.62500,0.39060,0.24875}%
\definecolor{mycolor5}{rgb}{0.78125,0.48825,0.31094}%
\definecolor{mycolor6}{rgb}{0.93750,0.58590,0.37312}%
\begin{tikzpicture}

\begin{axis}[%
width=0.27\textwidth,
height=1.5in,
at={(0.758in,0.481in)},
scale only axis,
xmin=0,
xmax=1,
xlabel style={font=\color{white!15!black}},
xlabel={$\beta$},
ymin=0.1,
ymax=1,
axis background/.style={fill=white}
]
\addplot [color=black, forget plot]
  table[row sep=crcr]{%
1	0.6068\\
0.95	0.6047\\
0.9	0.6037\\
0.85	0.601\\
0.8	0.5954\\
0.75	0.5843\\
0.7	0.5667\\
0.65	0.5447\\
0.6	0.531\\
0.55	0.5102\\
0.5	0.4279\\
0.45	0.3787\\
0.4	0.3517\\
0.35	0.3399\\
0.3	0.3325\\
0.25	0.3223\\
0.2	0.3157\\
0.15	0.3108\\
0.0999999999999997	0.3002\\
0.0499999999999997	0.2917\\
-3.19189119579733e-16	0.2648\\
};
\addplot [color=mycolor1, forget plot]
  table[row sep=crcr]{%
1	0.6018\\
0.95	0.5988\\
0.9	0.5958\\
0.85	0.5893\\
0.8	0.5809\\
0.75	0.5605\\
0.7	0.5419\\
0.65	0.5277\\
0.6	0.5164\\
0.55	0.5077\\
0.5	0.4835\\
0.45	0.4741\\
0.4	0.4638\\
0.35	0.4578\\
0.3	0.4572\\
0.25	0.447\\
0.2	0.4342\\
0.15	0.4181\\
0.0999999999999997	0.4072\\
0.0499999999999997	0.3909\\
-3.19189119579733e-16	0.369\\
};
\addplot [color=mycolor2, forget plot]
  table[row sep=crcr]{%
1	0.6004\\
0.95	0.6001\\
0.9	0.5994\\
0.85	0.6018\\
0.8	0.5982\\
0.75	0.5976\\
0.7	0.5969\\
0.65	0.5969\\
0.6	0.5955\\
0.55	0.5931\\
0.5	0.5911\\
0.45	0.5902\\
0.4	0.5884\\
0.35	0.5859\\
0.3	0.5818\\
0.25	0.577\\
0.2	0.5719\\
0.15	0.5655\\
0.0999999999999997	0.562\\
0.0499999999999997	0.5583\\
-3.19189119579733e-16	0.5082\\
};
\addplot [color=mycolor3, forget plot]
  table[row sep=crcr]{%
1	0.6218\\
0.95	0.621\\
0.9	0.6217\\
0.85	0.62\\
0.8	0.6193\\
0.75	0.6196\\
0.7	0.6187\\
0.65	0.6181\\
0.6	0.6169\\
0.55	0.618\\
0.5	0.6187\\
0.45	0.6195\\
0.4	0.6195\\
0.35	0.6191\\
0.3	0.6182\\
0.25	0.6182\\
0.2	0.6174\\
0.15	0.6163\\
0.0999999999999997	0.6142\\
0.0499999999999997	0.6131\\
-3.19189119579733e-16	0.6097\\
};
\addplot [color=mycolor4, forget plot]
  table[row sep=crcr]{%
1	0.5848\\
0.95	0.5873\\
0.9	0.5863\\
0.85	0.5846\\
0.8	0.5849\\
0.75	0.5856\\
0.7	0.5854\\
0.65	0.5851\\
0.6	0.585\\
0.55	0.5857\\
0.5	0.5854\\
0.45	0.5855\\
0.4	0.5856\\
0.35	0.5854\\
0.3	0.584\\
0.25	0.5831\\
0.2	0.5823\\
0.15	0.5805\\
0.0999999999999997	0.5798\\
0.0499999999999997	0.5795\\
-3.19189119579733e-16	0.5785\\
};
\addplot [color=mycolor5, forget plot]
  table[row sep=crcr]{%
1	0.6073\\
0.95	0.6084\\
0.9	0.6079\\
0.85	0.6059\\
0.8	0.6055\\
0.75	0.6064\\
0.7	0.605\\
0.65	0.6046\\
0.6	0.6045\\
0.55	0.6048\\
0.5	0.6051\\
0.45	0.6048\\
0.4	0.6044\\
0.35	0.6039\\
0.3	0.6035\\
0.25	0.6028\\
0.2	0.6024\\
0.15	0.6017\\
0.0999999999999997	0.6012\\
0.0499999999999997	0.6005\\
-3.19189119579733e-16	0.5997\\
};
\addplot [color=mycolor6, forget plot]
  table[row sep=crcr]{%
1	0.617733333333333\\
0.95	0.616933333333333\\
0.9	0.61624\\
0.85	0.61552\\
0.8	0.611946666666667\\
0.75	0.612106666666667\\
0.7	0.611653333333333\\
0.65	0.610853333333333\\
0.6	0.611653333333333\\
0.55	0.611706666666667\\
0.5	0.611653333333333\\
0.45	0.611786666666667\\
0.4	0.611226666666667\\
0.35	0.61152\\
0.3	0.611866666666667\\
0.25	0.61128\\
0.2	0.61096\\
0.15	0.610346666666667\\
0.0999999999999997	0.60904\\
0.0499999999999997	0.607813333333333\\
-3.19189119579733e-16	0.607066666666667\\
};
\addplot [color=mycolor6, draw=none, mark=asterisk, mark options={solid, mycolor6}, forget plot]
  table[row sep=crcr]{%
0.5	0.87\\
};
\addplot [color=black!30!mycolor5, draw=none, mark=+, mark options={solid, black!30!mycolor5}, forget plot]
  table[row sep=crcr]{%
0.5	0.59\\
};
\addplot [color=black, draw=none, mark=o, mark options={solid, black}, forget plot]
  table[row sep=crcr]{%
1	0.5\\
};
\end{axis}
\end{tikzpicture}%
} 
   \subfigure[Multi10\label{fig:resng20:multi10}]{
%
%
\definecolor{mycolor1}{rgb}{0.15625,0.09765,0.06219}%
\definecolor{mycolor2}{rgb}{0.31250,0.19530,0.12437}%
\definecolor{mycolor3}{rgb}{0.46875,0.29295,0.18656}%
\definecolor{mycolor4}{rgb}{0.62500,0.39060,0.24875}%
\definecolor{mycolor5}{rgb}{0.78125,0.48825,0.31094}%
\definecolor{mycolor6}{rgb}{0.93750,0.58590,0.37312}%
\begin{tikzpicture}

\begin{axis}[%
width=0.27\textwidth,
height=1.5in,
at={(0.758in,0.481in)},
scale only axis,
xmin=0,
xmax=1,
xlabel style={font=\color{white!15!black}},
xlabel={$\beta$},
ymin=0.1,
ymax=1,
axis background/.style={fill=white}
]
\addplot [color=black, forget plot]
  table[row sep=crcr]{%
1	0.6124\\
0.95	0.6176\\
0.9	0.6167\\
0.85	0.6164\\
0.8	0.6161\\
0.75	0.6154\\
0.7	0.6146\\
0.65	0.6129\\
0.6	0.6047\\
0.55	0.5894\\
0.5	0.2309\\
0.45	0.2065\\
0.4	0.1978\\
0.35	0.1921\\
0.3	0.1869\\
0.25	0.184\\
0.2	0.1832\\
0.15	0.1815\\
0.0999999999999997	0.1804\\
0.0499999999999997	0.1798\\
-3.19189119579733e-16	0.1771\\
};
\addplot [color=mycolor1, forget plot]
  table[row sep=crcr]{%
1	0.6039\\
0.95	0.6085\\
0.9	0.6147\\
0.85	0.6182\\
0.8	0.6209\\
0.75	0.6205\\
0.7	0.6181\\
0.65	0.611\\
0.6	0.5991\\
0.55	0.5474\\
0.5	0.3975\\
0.45	0.3419\\
0.4	0.3062\\
0.35	0.2794\\
0.3	0.2697\\
0.25	0.2533\\
0.2	0.2385\\
0.15	0.2351\\
0.0999999999999997	0.2426\\
0.0499999999999997	0.2465\\
-3.19189119579733e-16	0.2471\\
};
\addplot [color=mycolor2, forget plot]
  table[row sep=crcr]{%
1	0.6033\\
0.95	0.6038\\
0.9	0.6096\\
0.85	0.6133\\
0.8	0.6208\\
0.75	0.6225\\
0.7	0.6209\\
0.65	0.621\\
0.6	0.6176\\
0.55	0.5956\\
0.5	0.519\\
0.45	0.493\\
0.4	0.476\\
0.35	0.4623\\
0.3	0.4511\\
0.25	0.4372\\
0.2	0.4269\\
0.15	0.421\\
0.0999999999999997	0.4152\\
0.0499999999999997	0.4066\\
-3.19189119579733e-16	0.3746\\
};
\addplot [color=mycolor3, forget plot]
  table[row sep=crcr]{%
1	0.6145\\
0.95	0.6204\\
0.9	0.6313\\
0.85	0.6393\\
0.8	0.6404\\
0.75	0.6408\\
0.7	0.6401\\
0.65	0.6375\\
0.6	0.636\\
0.55	0.6349\\
0.5	0.6324\\
0.45	0.6295\\
0.4	0.6264\\
0.35	0.6233\\
0.3	0.6189\\
0.25	0.6123\\
0.2	0.6039\\
0.15	0.5931\\
0.0999999999999997	0.5815\\
0.0499999999999997	0.5467\\
-3.19189119579733e-16	0.5227\\
};
\addplot [color=mycolor4, forget plot]
  table[row sep=crcr]{%
1	0.615\\
0.95	0.6248\\
0.9	0.632\\
0.85	0.6425\\
0.8	0.6493\\
0.75	0.6519\\
0.7	0.6499\\
0.65	0.6496\\
0.6	0.6491\\
0.55	0.6471\\
0.5	0.6459\\
0.45	0.646\\
0.4	0.6455\\
0.35	0.6442\\
0.3	0.6438\\
0.25	0.6407\\
0.2	0.6382\\
0.15	0.6341\\
0.0999999999999997	0.6294\\
0.0499999999999997	0.6254\\
-3.19189119579733e-16	0.6181\\
};
\addplot [color=mycolor5, forget plot]
  table[row sep=crcr]{%
1	0.6095\\
0.95	0.6179\\
0.9	0.6262\\
0.85	0.6338\\
0.8	0.6375\\
0.75	0.6388\\
0.7	0.6409\\
0.65	0.6407\\
0.6	0.6393\\
0.55	0.6384\\
0.5	0.637\\
0.45	0.6356\\
0.4	0.6345\\
0.35	0.634\\
0.3	0.6333\\
0.25	0.632\\
0.2	0.6309\\
0.15	0.6289\\
0.0999999999999997	0.6275\\
0.0499999999999997	0.6217\\
-3.19189119579733e-16	0.618\\
};
\addplot [color=mycolor6, forget plot]
  table[row sep=crcr]{%
1	0.6184\\
0.95	0.6216\\
0.9	0.6241\\
0.85	0.6255\\
0.8	0.626\\
0.75	0.6287\\
0.7	0.6286\\
0.65	0.6289\\
0.6	0.6291\\
0.55	0.6295\\
0.5	0.6298\\
0.45	0.6293\\
0.4	0.6282\\
0.35	0.6282\\
0.3	0.6269\\
0.25	0.6257\\
0.2	0.6232\\
0.15	0.6214\\
0.0999999999999997	0.619\\
0.0499999999999997	0.6159\\
-3.19189119579733e-16	0.6103\\
};
\addplot [color=mycolor6, draw=none, mark=asterisk, mark options={solid, mycolor6}, forget plot]
  table[row sep=crcr]{%
0.5	0.56\\
};
\addplot [color=black!30!mycolor5, draw=none, mark=+, mark options={solid, black!30!mycolor5}, forget plot]
  table[row sep=crcr]{%
0.5	0.35\\
};
\addplot [color=black, draw=none, mark=o, mark options={solid, black}, forget plot]
  table[row sep=crcr]{%
1	0.29\\
};
\end{axis}
\end{tikzpicture}%
} 
   \caption{Micro-averaged precision for different NG20 subsets and \textsc{AnnITCC}. Results are shown for different numbers of word clusters, $|\overline{\mathcal{W}}| = \{2,4,8,16, 32, 64, 128 \}$ (darker colors for fewer clusters). For comparison, we added results reported in the literature. $(*)$: Taken from~\cite[Table~5]{Dhillon_ITClustering}; $|\overline{\mathcal{W}}|$ is unclear. $(+,\circ)$: Taken from~\cite[Table~3]{Slonim_DoubleClustering}; the best results for each dataset are displayed. These results were obtained by applying aIB for different numbers of word clusters, $|\overline{\mathcal{W}}| = \{10,20,30,40,50\}$; the displayed MAP values are averages of the individual MAP values. We were not able to compare our results to those of~\cite{Wang_IBCC} because they used different subsets of the NG20 dataset. Since the cost functions from the literature are the same as ours for the respective values of $\beta$, the difference in the performance can only be attributed to preprocessing steps, the optimization heuristics, and/or the choice of favorable data subsets.}
    \label{resng20} 
\end{figure*}

The Newsgroup20 (NG20) dataset\footnote{qwone.com/$\sim$jason/20Newsgroups} consists of approximately 18800 documents containing 50000 different words. In this section, we evaluate co-clustering performance only via document clusters since there is no ground truth for word clusters. Nevertheless, word clustering was claimed to improve the document clustering performance, cf.~\cite{Slonim_DoubleClustering,Dhillon_ITClustering}. 

We refer to the RV over words as $W$, the set of words as $\mathcal{W}$, the RV over the documents as $D$, and the set of documents as $\mathcal{D}$. The respective clustered RVs and sets are denoted by an overline. The joint distribution of $W$ and $D$ is obtained by normalizing the contingency table (counting the number of times a word appears in a document) to a probability distribution. During preprocessing, we removed newsgroup-identifying headers and lowered upper-case letters. We moreover reduced $\mathcal{W}$ to the 2000 words with the highest contribution to $I(D;W)$, which is consistent with the preprocessing in~\cite{Slonim_DoubleClustering,Slonim_DocClustering,Dhillon_ITClustering}. Finally, we constructed various subsets of the NG20 dataset by randomly selecting 500 documents evenly distributed among the document classes. An overview of the used datasets is given in Table~\ref{20ng1}. 

Note that there are significant differences in the preprocessing steps performed in previous studies. For example,~\cite{Slonim_DocClustering} included the newsgroup-identifying header, which may improve clustering performance.

\begin{table}[t]
\newcolumntype{P}[1]{>{\centering\arraybackslash}p{#1}}
\centering   
    \caption{Overview of the different subsets drawn from NG20}
    \label{20ng1}
    \begin{tabular}{ | c || P{5cm} | c | c |}
    \hline 
    Dataset & Discussion Groups  & $\frac{\mathrm{docs}}{\mathrm{class}}$ & $|\mathcal{D}|$ \\ \hline \hline 
    Binary  & talk.politics.mideast, talk.politics.misc & 250 & 500 \\ \hline
    Multi5  & rec.motorcycles, comp.graphics, sci.space, rec.sport.basketball, talk.politics.mideast & 100 & 500  \\ \hline
    Multi10 & comp.sys.mac.hardware, misc.forsale, rec.autos, talks.politics.gun, sci.med, alt.atheism, sci.crypt, sci.space, sci.electronics, rec.sport.hockey & 50  & 500 \\ \hline 
    \end{tabular} 
\end{table}

We ran \textsc{AnnITCC} with $\text{tol} = 10^{-3}$, $\Delta = 0.05$ and $\#\text{iter}_{\text{max}} = 20$. For initialization, we slightly changed line 3 in Algorithm~\ref{betaCC}: Instead of running \textsc{sGITCC} with $\beta = 1$, which is equivalent to the completely decoupled case, we run sIB for both the word and document clusterings separately, where 25 restarts are performed and the best result w.r.t.\ the cost function is taken. Since there is no ground truth available for the word clusters, we executed \textsc{AnnITCC} for $|\overline{\mathcal{W}}| \in\{2,4,8,16, 32, 64, 128 \}$. This is consistent with the simulation settings described in \cite{Dhillon_ITClustering}, for example. 

For a fair comparison of different values of $\beta$, we do not apply further heuristics to improve the performance of \textsc{AnnITCC}. In contrast, the authors of \cite{Dhillon_ITClustering} initialize their co-clustering algorithm for $|\overline{\mathcal{W}}|$ word clusters with the result obtained for $|\overline{\mathcal{W}}|/2$ word clusters, where each word cluster is split randomly. In \cite{Bekkerman_MultiClustering}, the authors introduce an additional correction parameter which leads to clusters of approximately the same size (which matches the evenly distributed classes in the NG20 dataset). Therefore, even for those values of $\beta$ for which we obtain the same cost functions, our results need not be equal to those reported in the literature.

\subsubsection{Results and Comparison}

The results obtained by Algorithm \ref{betaCC} - averaged over 20 runs - for the different subsets of NG20 are visualized in Fig.~\ref{resng20}. As it can be seen, \textsc{AnnITCC} can discover the true document labels with high accuracy. For the Binary dataset, \textsc{AnnITCC} was able to achieve a micro-averaged precision of approximately $90\%$, for the Multi5 dataset $60 \%$ and for the Multi10 dataset approximately $60-65 \%$. In comparison, experiments with \textsc{sGITCC} confirm the observations from \cite{Amjad_GeneralizedMA} that small $\beta \in [0, 0.4]$ lead to meaningless results in the range of random clustering, while high $\beta \in [0.6, 1]$ produce results in the range of Fig.~\ref{resng20}. Fig.~\ref{resng20} further shows that the \rev{stronger the word and document clustering solutions are coupled}, the worse are the results for small numbers of word clusters. \rev{This is most obvious for the Multi10 dataset for $\overline{\mathcal{W}}\in\{2,4,8\}$ word clusters, where the MAP values increase sharply if $\beta$ increases from 0.4 to 0.6 (see Fig.~\ref{fig:resng20:multi10}). For small $\beta$, the document clusters are obtained from the word clusters and, e.g., two word clusters do not contain sufficient information to distinguish between ten document clusters.} This agrees with our discussion in Section~\ref{sec:differentCardinalities}. However, for very large $|\overline{\mathcal{W}}|$, there were no further improvements. This suggests that there exists a number of word clusters that are sufficient to achieve the same (or better, see below) performance as document clustering based on words.

One major issue to observe from Fig.~\ref{resng20} is that for the Binary and Multi5 data, the results are almost independent of $\beta$ (for sufficiently many word clusters). Only for Multi10 there was a mild increase in performance for intermediate values of $\beta$. This confirms the observations from Section~\ref{sec:synthetic}: Clustering words removes noise, hence document clustering based on word clusters may be slightly more robust than document clustering based on words. Nevertheless, since the effect is only small for Multi10 (and not present for Binary and Multi5), we doubt that co-clustering of words and documents is indeed significantly superior to one-sided document clustering w.r.t.\ the classification results. The classification results from \cite{Bekkerman_MultiClustering} point towards similar conclusions, since also there sIB performed very well compared to the respective co-clustering methods. Still, the authors of~\cite{Slonim_DoubleClustering,Dhillon_ITClustering,Wang_IBCC} claim that their proposed algorithms and/or cost functions for co-clustering outperform one-sided clustering. In the light of our results, we suggest that the choice of the cost function has less effect on the performance than algorithmic details, preprocessing steps, and additional heuristics for, e.g., initialization.

\subsection{MovieLens100k}
\subsubsection{Dataset, Preprocessing, and Simulation Settings}

The MovieLens100k dataset\footnote{grouplens.org/datasets/movielens/100k} consists of $100000$ ratings of $1682$ movies by $943$ users~\cite{movielens}. The user ratings take integer values $1$ (worst) to $5$ (best). We construct a user-movie matrix $\Rvec:=[R_{ij}]$ where $R_{ij}$ is the rating user $i$ gave to the movie $j$ ($R_{ij}=0$ if user $i$ did not rate movie $j$). Note that $\Rvec$ is a sparse matrix with only $100000$ out of approximately $1.59$ million entries being nonzero.

We refer to the RV over the users as $U$, the set of users as $\mathcal{U}$, the RV over movies as $M$, and the set of movies as $\mathcal{M}$. The respective clustered RVs and sets are denoted by an overline. The joint distribution between $U$ and $M$ is obtained by normalizing $\Rvec$ to a probability distribution.

For initializing \textsc{AnnITCC} we ran \textsc{sGITCC} $25$ times with random initializations for $\beta =1$ with $\text{tol} = 10^{-3}$ and $\#\text{iter}_{\text{max}} = 20$. We chose the best co-clustering $(\Phi, \Psi)$ among these $25$ restarts w.r.t. the cost and used this as the initialization for \textsc{AnnITCC}. We ran \textsc{AnnITCC} with $\text{tol} = 10^{-3}$, $\Delta = 0.1$ and $\#\text{iter}_{\text{max}} = 20$. We defined $10$ user clusters, i.e., $|\overline{\mathcal{U}}| = 10$, as was done in \cite{Banerjee_Bregman,Laclau_OptimalTransport}. Furthermore, we defined $|\overline{\mathcal{M}}| = 19$ since the MovieLens100k dataset categorizes the movies into $19$ different genres.

\subsubsection{Evaluation Metrics}

Evaluating co-clustering performance for the MovieLens100k dataset is difficult. The authors of~\cite{Laclau_OptimalTransport} proposed to assess co-clustering performance based on recommendations, i.e., a portion of the dataset is used for co-clustering, based on which the ``taste'' of the users is predicted. The remaining portion of the dataset (i.e., the validation set) is used to assess this prediction. We believe that such an approach is not effective. Indeed, the available ratings in $\Rvec$ are skewed in the sense that approximately $82.5\%$ of the ratings are above $3$. Hence, a naive recommendation system suggesting a positive rating for every user-movie pair in the validation set matches the user's taste with approximately $82.5\%$. In comparison, the authors of~\cite{Laclau_OptimalTransport} claim a match of $89\%$ for their approach. 

A second option is to compare the co-clustering results to a plausible ground truth. For the users, demographic information is available which theoretically admits constructing such a ground truth; we nevertheless refrain from doing so, since no choice can be justified without evoking critique. For the movies, genre information is available which lends itself to evaluating movie clusters. However, not every movie is assigned to a unique genre, but may belong to multiple genres. The ground truth $\Psi^\bullet$ is therefore not a function, but a distribution over the set of genres $\overline{\mathcal{M}}$. This is problematic for~\eqref{eq:map}, which is why we replace it here by
\begin{equation}
\mathrm{MAP'}(\Psi,\Psi^\bullet) :=  \frac{1}{|\dom{M}|}\sum_{j\in\overline{\dom{M}}} \max\limits_{i\in\overline{\dom{M}}}|\Psi^{-1}(j)\cap \Psi^{\bullet-1}(i)| .
\end{equation}
For each movie cluster, we look for the genre with which this cluster has the greatest overlap. Unlike for $\mathrm{MAP}$, two different clusters can now be mapped to same movie genre in $\mathrm{MAP'}$. Hence, $\mathrm{MAP'}$, sometimes referred to as \emph{purity}, is essentially the average of the fraction of movies in each cluster that belong to the same genre. As a side result, $\mathrm{MAP'}$ gets rid of the maximum over all permutations $\pi$, which is intractable for large numbers of genres.

\subsubsection{Results}

\begin{figure}
	\centering
\begin{tikzpicture}
\begin{axis}[
width=0.45\textwidth,
height=1.75in,
ylabel={$\mathrm{MAP'}(\Psi,\Psi^\bullet)$},
xlabel={$\beta$},
xmin=0, xmax=1,
ymin=0.4, ymax=0.6,
minor y tick num={1},
minor x tick num={1},
legend style={at={(0.03,0.97)},anchor=north west},
]
\addplot[blue, line width=2.0pt] file {genrematchtex/genrematch.txt};
\addlegendentry{\footnotesize \textsc{AnnITCC}}
\addplot[mark=none, red, line width=2.0pt] {0.431};
\addlegendentry{\footnotesize Random Movie Clusters}
\end{axis}
\end{tikzpicture}
	  \caption{\textsc{AnnITCC} performance for movie genre matching}
	  \label{fig:movielens1}
\end{figure}
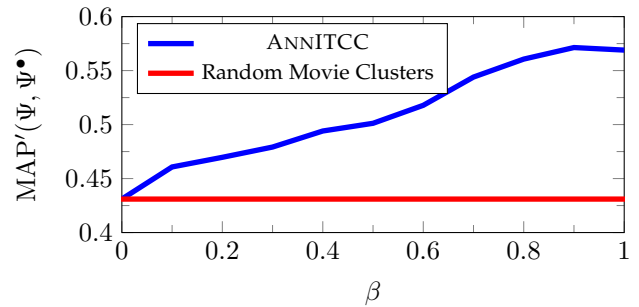

The results are shown in Fig.~\ref{fig:movielens1}. First, note that the $\textsc{MAP'}$ value for randomly generated clusters is remarkably high. This is because the number of movies in different genres varies greatly; for example, $725$ movies are assigned to genre \enquote{Drama} and $505$ to genre \enquote{Comedy}, whereas only $24$ movies belong to the genre \enquote{Film-Noir}. Noting this, quantitative results based on movie genres are useful to observe trends and general behavior, but the numbers should be taken with a grain of salt. On the other extreme, the maximum value for $\mathrm{MAP'}$ in Fig.~\ref{fig:movielens1} is significantly smaller than $1$. This is reasonable since co-clustering is based on a sparse matrix of user-movie rating pairs: While some users are \emph{genre-addicts} rating movies mainly based on their genre, other users may rate movies based on completely different aspects unrelated to genre. Hence, one cannot expect a value $\mathrm{MAP'}=1$ for co-clustering based on user-movie rating pairs. 

We observe that $\mathrm{MAP'}$ generally decreases with decreasing $\beta$ and the maximum value is at $\beta=0.9$, albeit only slightly larger than for $\beta=1$. This shows that our algorithm is capable of outperforming ITCC ($\beta=\frac{1}{2}$), IBCC ($\beta=\frac{3}{4}$), and (albeit only slightly) IB-based ($\beta=1$) movie clustering. For $\beta$ close to $0$, we obtain results which are very close to what we obtain for randomly generated movie clusters. A closer analysis revealed that the solution found for $\beta=0$ has a lower cost than the solution found for $\beta=1$, which means that $\beta$-annealing was successful in escaping bad local optima, but that the ground truth does not coincide with the global optimum of the cost function for $\beta=0$. We believe that, in this particular example, this phenomenon is linked to the user-movie rating matrix $\mathbf{R}$ being sparse.

We finally complement this quantitative evaluation by a qualitative evaluation of the movie clusters. Again, we observe meaningful results for higher values of $\beta$ when compared to smaller values of $\beta$. For example, looking at movie clusters for $\beta=0.9$, we notice that many classics are clustered into one group, including \textit{Gone With The Wind}, \textit{Breakfast at Tiffany's (1961)}, \textit{12 Angry Men}, \textit{The Graduate}, \textit{The Bridge on River Kwai}, \textit{Citizen Kane}, \textit{Dr. Strangelove or: How I Learned to Stop Worrying and Love the Bomb}, \textit{Vertigo}, \textit{Casablanca}, \textit{His Girl Friday (1940)}, \textit{A Street Car Named Desire}, \textit{It Happened One Night}, \textit{The Great Dictator}, \textit{The Great Escape}, \textit{Philadelphia Story}. Similarly, many animated/kids movies have been assigned to a cluster, including \textit{The Lion King}, \textit{Alladin},  \textit{Snow White and the Seven Dwarfs}, \textit{Homeward Bound}, \textit{Pinocchio}, \textit{Turbo: A Power Rangers Movie}, \textit{Mighty Morphin Power Rangers: The Movie}, \textit{Cinderella}, \textit{Alice in Wonderland (1951)}, \textit{Dumbo (1941)}, \textit{Beauty and the Beast}, \textit{Winnie the Pooh and the Blustery Day}, \textit{The Jungle Book}, \textit{The Fox and the Hound}, \textit{Parent Trap}, \textit{Jumanji}, \textit{Casper}, etc. Furthermore, our approach clustered various sequences of movies, e.g., $6$ out of $8$ Star Trek movies and all $7$ Amityville movies have been assigned to one cluster each. In contrast, the results for $\beta=0$ did not yield clusters one would consider meaningful.

\subsection{Community Detection in Bipartite Graphs}\label{sec:community}
\label{CD}

Community detection is a common problem in social network analysis and is usually concerned with (random) unipartite graphs, see~\cite{Fortunato_CDGraphs}. In this section, we look at the related problem for bipartite graphs. There, the two sets of vertices could be the characters and the scenes of a play, and the goal could be to group characters in a meaningful way. 

\begin{figure}[t]
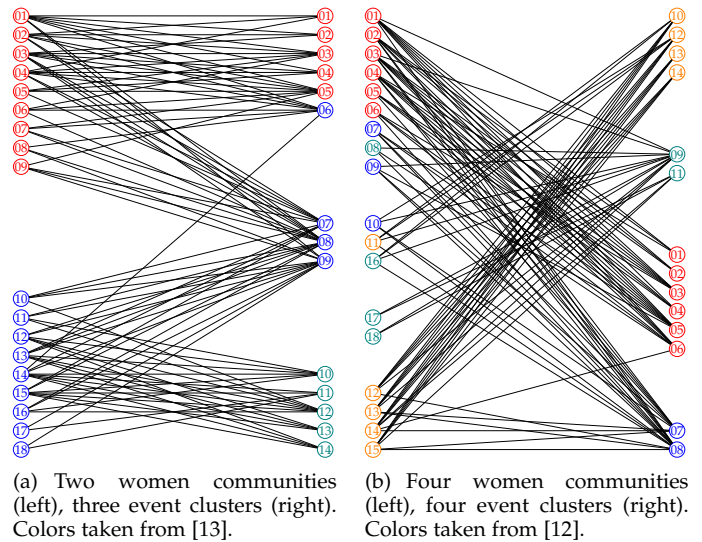

\centering 
 \subfigure[Two women communities (left), three event clusters (right). Colors taken from~\cite{Doreian_Blockmodeling}.]{\scalebox{0.5}{\begin{tikzpicture}
	
	\input{GroundTruth10}
	\input{paths}
\end{tikzpicture}}\label{fig:sw:23}}\hfill
 \subfigure[Four women communities (left), four event clusters (right). Colors taken from~\cite{Barber_Modularity}.]{\scalebox{0.5}{\begin{tikzpicture}
	\input{Barber4407}
	\input{paths}
\end{tikzpicture}}\label{fig:sw:44}}
    \caption{Community Structure of the Southern Women Event Participation Dataset. The separation between nodes indicates the clustering obtained from \textsc{AnnITCC} with $\beta=0.7$, the color of the nodes is taken from reference clusterings from the literature.}
    \label{fig:sw} 
\end{figure}

We apply our algorithm to the Southern Women Event Participation Dataset \cite{Barber_Modularity,Fortunato_CDGraphs}. The dataset consists of 18 women ($|\dom{X}|=18$) and 14 events ($|\dom{Y}|=14$), and the weight matrix $\Wvec$ contains a one if the corresponding woman participated in the corresponding event and a zero otherwise. We restarted \textsc{AnnITCC} 50 times for $\beta=1$ to obtain a good initial co-clustering for the annealing process. To get results comparable to those in the literature, we chose $|\overline{\dom{X}}| = 2, |\overline{\dom{Y}}| = 3$ and $|\overline{\dom{X}}| = |\overline{\dom{Y}}| = 4$. The results are displayed in Fig.~\ref{fig:sw} for $\beta =0.7$. 

The two women communities we obtained match with those communities reported in the literature~\cite{Fortunato_CDGraphs,Doreian_Blockmodeling}. The authors of~\cite{Doreian_Blockmodeling} also clustered the events into three clusters: The events are clustered into a group in which only women of the first women community participated, a group in which only women of the second women community participated, and a group in which women from both communities participated. Our result in Fig.~\ref{fig:sw:23} is remarkably similar to theirs, with the exception that the event with label 6 is put in a different group. Note, however, that in this event only one woman of the opposite community participated. Remarkably, we obtained the same co-clustering for all values of $\beta$.

For four women communities and four event clusters, we compared our results with those of Barber~\cite{Barber_Modularity}, who employed a modularity-based approach. Our event clusters in Fig.~\ref{fig:sw:44} are identical to those of~\cite{Barber_Modularity}, and our women communities are largely consistent. We found in a separate set of experiments that the women communities show a greater agreement for $\beta=1$, and less agreement for $\beta=\frac{1}{2}$; the MAP values for the chosen value of $\beta=0.7$ lie in between. Thus, community detection via ITCC can be outperformed by our algorithm for larger values of $\beta$.

\section{Conclusion}

We introduced a generalized framework for information-theoretic co-clustering that arises from recent results on the theory of Markov aggregation. The generalized cost function we proposed allows for trading between completely coupled and decoupled clusterings of two variables connected via a probability table. We obtain well-known previous approaches, e.g., Information-Theoretic Co-Clustering from Dhillon et al., as special cases of our cost function. Using this framework, we provided better understanding of information-theoretic co-clustering in general and discussed some shortcomings inherent to such co-clustering methods. 

We performed experiments on both synthetic and real-world data, such as document classification, movie clustering, and community detection. We also demonstrated that our framework can be used to fairly compare various previously proposed cost functions. For example, for the Newsgroup20 dataset, we observed that performance depended little on the cost function, but rather on the optimization heuristic, preprocessing steps, and/or choice of data subsets. \rev{We furthermore provide guiding principles for choosing the parameter $\beta$ of our cost function depending upon the characteristics of the dataset.}

\section*{Acknowledgments}
The work of Rana Ali Amjad has has been funded by the German Ministry of Education and Research in the framework of an Alexander von Humboldt Professorship.
The work of Bernhard C. Geiger has been funded by the Erwin Schr\"odinger Fellowship J 3765 of the Austrian Science Fund.

\bibliographystyle{IEEEtran}
\bibliography{IEEEabrv,references}

\begin{IEEEbiography}[{\includegraphics[width=1in,height=1.25in,clip,keepaspectratio]{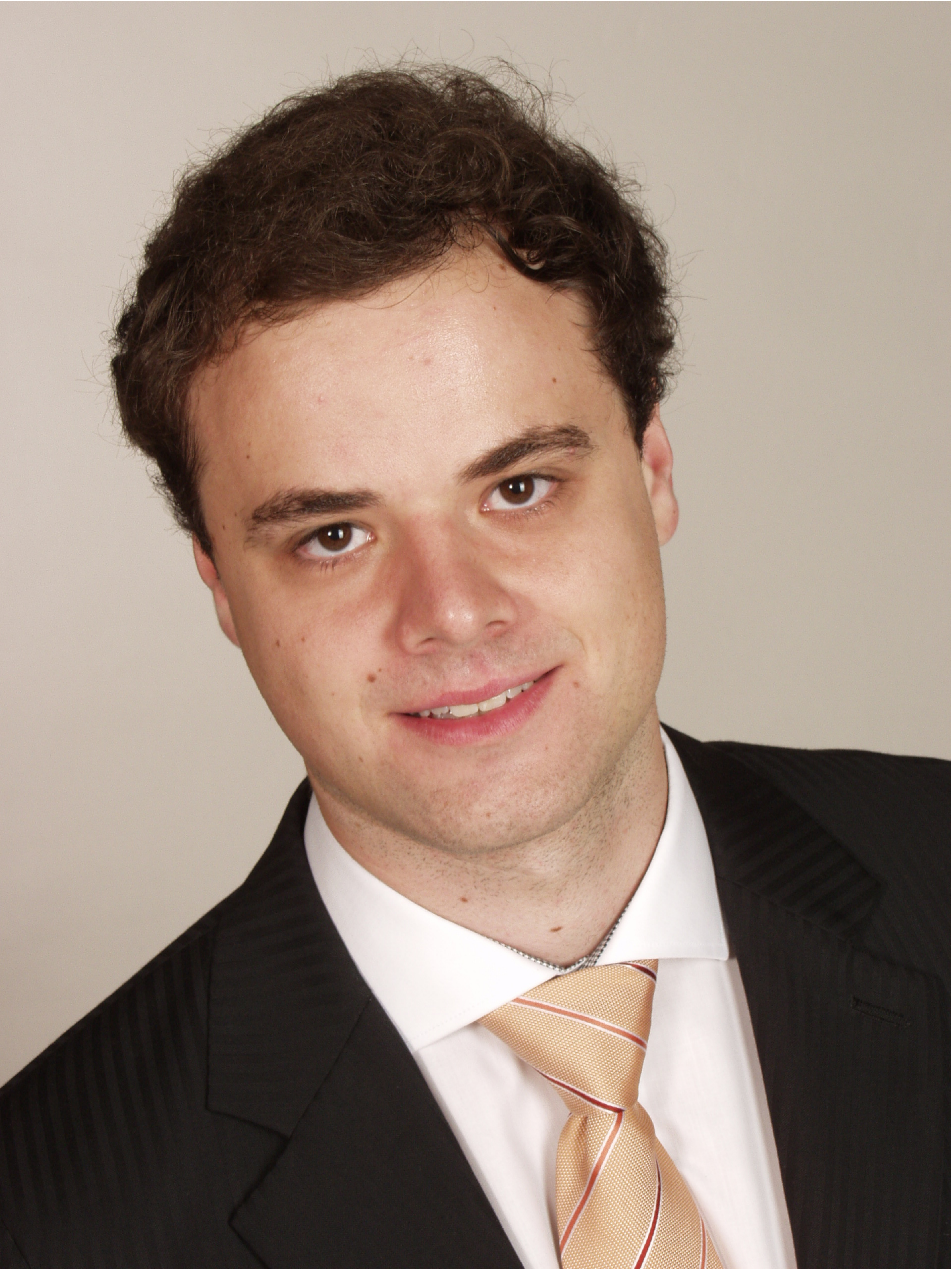}}]{Clemens Bl\"ochl}
was born in Traunstein, Germany, in 1992. He received his B.Sc. and M.Sc. degree from Technical University of Munich, Germany, in 2014 and 2017, respectively, all in electrical engineering and information technology. 

In 2017 he joined Rohde \& Schwarz GmbH \& Co. KG in Munich, where he is currently involved in the field of radio communication testing. His research interests include channel coding theory and information-theoretic system design with particular interest in the field of information-theoretic clustering.
\end{IEEEbiography}

\begin{IEEEbiography}[{\includegraphics[width=1in,height=1.25in,clip,keepaspectratio]{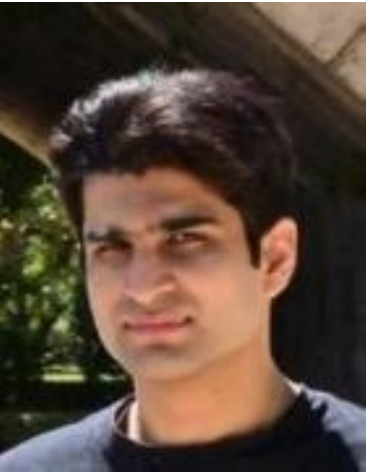}}]{Rana Ali Amjad}
	(S'13) was born in Sahiwal, Pakistan, in 1989. He received the Bachelors degree in Electrical Engineering  (with highest distinction) from University of Engineering and Technology, Lahore, Pakistan, in 2011. He completed his Masters degree in Communication Engineering (with highest distinction) from Technical University of Munich, Germany, in 2013.  
	
	Since 2014 he is pursuing his PhD at the Institute for Communication Engineering at Technical University of Munich. He has received various awards in his academic career including the faculty award for best Master thesis, award for outstanding performance in Master's degree and Gold medal for best performance in Communications major during his Bachelors degree. His research interests cover information and communication theory, machine learning, channel coding and information-theoretic security.
\end{IEEEbiography}

\begin{IEEEbiography}[{\includegraphics[width=1in,height=1.25in,clip,keepaspectratio]{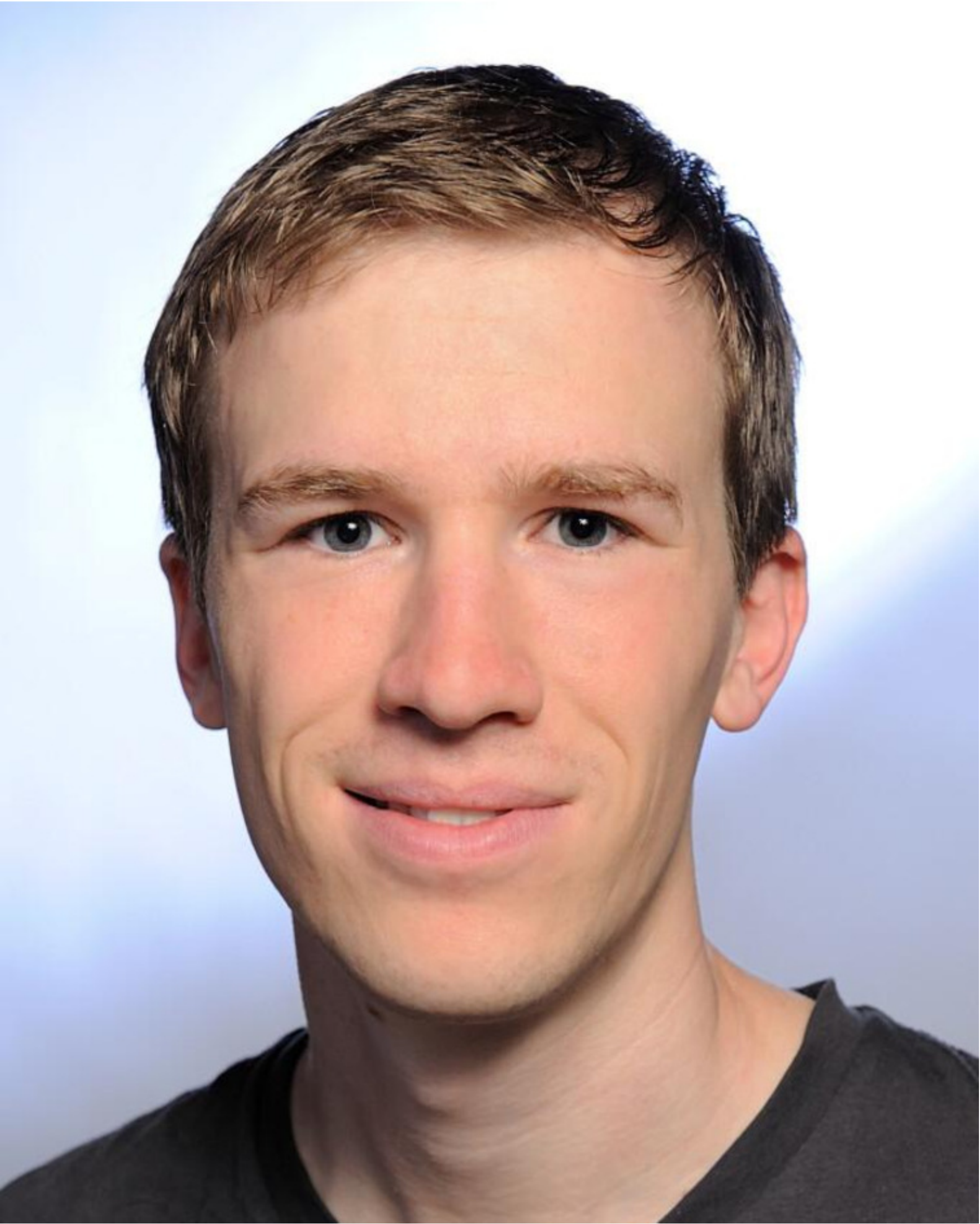}}]{Bernhard C. Geiger}
(S'07, M'14) was born in Graz, Austria, in 1984. He received the Dipl.-Ing. degree in electrical engineering (with distinction) and the Dr. techn. degree in electrical and information engineering (with distinction) from Graz University of Technology, Austria, in 2009 and 2014, respectively.

In 2009 he joined the Signal Processing and Speech Communication Laboratory, Graz University of Technology, as a Project Assistant and took a position as a Research and Teaching Associate at the same lab in 2010. He was a Senior Scientist and Erwin Schr\"odinger Fellow at the Institute for Communications Engineering, Technical University of Munich, Germany from 2014 to 2017 and a postdoctoral researcher at the Signal Processing and Speech Communication Laboratory, Graz University of Technology, Austria from 2017 to 2018. He is currently a Senior Researcher at KNOW-CENTER GmbH, Graz, Austria. His research interests cover information theory for machine learning and information-theoretic model reduction for Markov chains and hidden Markov models.
\end{IEEEbiography}

\end{document}